%% file: egpaper_final.tex
\documentclass[10pt,twocolumn,letterpaper]{article}

\usepackage{iccv}
\usepackage{times}
\usepackage{epsfig}
\usepackage{graphicx}
\usepackage{amsmath}
\usepackage{amssymb}
\usepackage{caption}
\usepackage{algorithm}
\usepackage{algpseudocode}
\usepackage{bbm}
\usepackage[caption=false]{subfig}

\usepackage[breaklinks=true,bookmarks=false]{hyperref}

\iccvfinalcopy 

\def\iccvPaperID{****} 

\DeclareMathOperator*{\argmax}{arg\,max}

\DeclareMathOperator*{\E}{\mathbb{E}}

\ificcvfinal\fi

\begin{document}

\title{Game Theoretic Mixed Experts for Combinational Adversarial Machine Learning}

\author{Ethan Rathbun, Kaleel Mahmood, Caiwen Ding, Sohaib Ahmad\\
University of Connecticut\\
{\tt\small \{first\}.\{last\}@uconn.edu}
\and
Marten Van Dijk\\
Centrum Wiskunde \& Informatica, Amsterdam\\
{\tt\small marten.van.dijk@cwi.nl}
}
\maketitle
\ificcvfinal\thispagestyle{empty}\fi
\begin{abstract}
Recent advances in adversarial machine learning have shown that defenses considered to be robust are actually susceptible to adversarial attacks which are specifically customized to target their weaknesses. These defenses include Barrage of Random Transforms (BaRT), Friendly Adversarial Training (FAT), Trash is Treasure (TiT) and ensemble models made up of Vision Transformers (ViTs), Big Transfer models and Spiking Neural Networks (SNNs). We first conduct a transferability analysis, to demonstrate the adversarial examples generated by customized attacks on one defense, are not often misclassified by another defense. 

This finding leads to two important questions. First, how can the low transferability between defenses be utilized in a game theoretic framework to improve the robustness? Second, how can an adversary within this framework develop effective multi-model attacks? In this paper, we provide a game-theoretic framework for ensemble adversarial attacks and defenses. Our framework is called Game theoretic Mixed Experts (GaME). It is designed to find the Mixed-Nash strategy for both a detector based and standard defender, when facing an attacker employing compositional adversarial attacks. We further propose three new attack algorithms, specifically designed to target defenses with randomized transformations, multi-model voting schemes, and adversarial detector architectures. These attacks serve to both strengthen defenses generated by the GaME framework and verify their robustness against unforeseen attacks. Overall, our framework and analyses advance the field of adversarial machine learning by yielding new insights into compositional attack and defense formulations.
\end{abstract}
\input{Sections/Introduction2}
\input{Sections/Defenses}
\input{Sections/Attacks}
\input{Sections/ExperimentsTransfer}
\input{Sections/Game}
\input{Sections/ExperimentsGame}
\input{Sections/Conclusion}
\newpage
{\small
\bibliographystyle{ieee_fullname}
\bibliography{egpaper_final}
}
\clearpage
\input{Sections/Appendix}

\end{document}

%% file: Sections/Introduction2.tex
\section{Introduction}

Machine learning models have been shown to be vulnerable to adversarial examples~\cite{goodfellow2014explaining, papernot2016transferability}. Adversarial examples are inputs with small perturbations added, such that machine learning models misclassify them with high confidence. Addressing the security risks posed by adversarial examples are critical for the safe deployment of machine learning in areas like health care~\cite{medicineAdvML} and self driving vehicles~\cite{drivingAdvML}. However, current defenses and attacks in adversarial machine learning have trended towards a cat and mouse dynamic, where in new defenses are being proposed and then broken~\cite{bypassing10detectors, tramer2020adaptive, mahmood2021beware, demystifying} by improved attacks.  

In parallel to attack and defense development, studies have also been conducted on the transferability of adversarial examples~\cite{Delving, mahmood2021robustness, xu2022securing}. Transferabiltiy refers to the phenomena where adversarial examples generated for one model are also misclassified by a different machine learning model. However, to the best of our knowledge no analyses have been done on the transferability of adversarial examples designed to attack specific defenses. From these observations several pertinent questions arise:
\begin{enumerate}
    \item \textit{Do adversarial examples generated for one specific defense transfer to other defenses?}
    \item \textit{Based on adversarial transferability, can a game theoretic framework be developed to determine the optimal choices for both attacker and defender?}
    \item \textit{How can the addition of multi-defense consensus voting (adversarial detection) and multi-model attacks be exploited by both the attacker and defender within a game theoretic framework?}
\end{enumerate}
These are precisely the questions our paper seeks to answer. We break from the traditional dynamic of adversarial machine learning, which focuses on the single best attack and defense. We instead take a multi-faceted approach and develop a game theoretic framework to answer the above questions. We provide the following contributions: 
\begin{enumerate}
    \item \textbf{Defense Transferability Analysis}- We analyze the adversarial transferability of state-of-the-art defenses like Trash is Treasure~\cite{TiT}, Barrage of Random Transforms~\cite{BaRT}, Friendly Adversarial Training~\cite{FAT} and other new architectures like SNNs~\cite{transferSNN, SEW} and ViTs~\cite{VIT}. We show specific attacks on defenses do no transfer well, meaning there is potential for a combinational defense within a game theoretic framework.
    \item \textbf{New Combinational Adversarial Attacks} - We develop three new attacks: the Momentum Iterative Method over Expectation attack (MIME), the Multi-Agent Gradient Expectation attack (MAGE) and detector based Multi-Agent Gradient Expectation attack (MAGE-D). These attacks are designed to create the strongest possible adversary to attack randomized and multi-classifier based defenses.  
    \item \textbf{Game theoretic Mixed Experts} - We propose a game-theoretic framework for finding approximately optimal strategies for adversarial attackers and defenders that can implement multi-model attack and defense techniques. We mathematically derive our framework, and empirical test it  by implementing multiple state-of-the-art adversarial attacks and defenses on two datasets, CIFAR-10~\cite{c10} and Tiny ImageNet~\cite{le2015tiny}. Through our framework we demonstrate that we are able to achieve a 37.3\% increase in robustness on CIFAR-10, and a 45.65\% increase on Tiny ImageNet over all single-model defenses we tested. 
\end{enumerate}



In contrast to previous works studying similar adversarial games~\cite{sengupta2018mtdeep}, our framework is more comprehensive, state-of-the-art, and includes broader analyses. Specifically, we consider an adversary that can employ both state-of-the-art single model and multi-model attacks (of which we formulate three new ones). Likewise, our defender is different from~\cite{sengupta2018mtdeep} in that they can utilize randomized transformations, different voting schemes, and adversarial detection mechanisms. Overall, our paper demonstrates that by using our framework, higher robustness can be achieved than any single state-of-the art defense we studied. 

%% file: Sections/Defenses.tex
\section{Adversarial Machine Learning Defenses}

In the following subsections, we give an overview of each defense and our reasons for choosing said defense. It is important to note our analyses encompass a broad range of different defenses, including ones based on randomization, ones based on adversarial training and ones based on exploiting model transferability. In addition, we also consider diverse architectures including Big Transfer models (BiTs), Vision Transformers (ViTs) and Spiking Neural Networks (SNNs). Despite our broad range, we do not attempt to test every novel adversarial defense. It is simply infeasible to test every proposed adversarial machine learning defense, as new defenses are constantly being produced. Corresponding code to replicate our results can be found \href{https://github.com/EthanRath/Game-Theoretic-Mixed-Experts}{here}.  

\subsection{Barrage of Random Transforms} 

Barrage of Random Transforms (BaRT)~\cite{BaRT} utilize a set of image transformations in a random order and with randomized transformation parameters to thwart adversarial attacks. In this paper, we work with the original BaRT implementation which includes both differentiable and non-differentiable image transformations.

    \label{eq:bart}

\textbf{Why we selected it:} Many defenses are broken soon after being proposed~\cite{tramer2020adaptive}. BaRT is one of the few defenses that has continued to show robustness even when attacks are specifically tailored to work against it. For example, most recently BaRT achieves $29\%$ robustness on CIFAR-10 against a customized white-box attack~\cite{demystifying}. It remains an open question whether using BaRT with other randomized approaches (i.e. selecting between different defenses) can yield even greater robustness.

\subsection{Friendly Adversarial Training} 

Training classifiers to correctly recognize adversarial examples was originally proposed in~\cite{goodfellow2014explaining} using FGSM. This concept was later expanded to include training on adversarial examples generated by PGD in~\cite{madry2018towards}. In~\cite{FAT}, it was shown that Friendly Adversarial Training (FAT) could achieve high clean accuracy, while maintaining robustness to adversarial examples. This training was accomplished by using a modified version of PGD called PGD-$K$-$\tau$. In PGD-$K$-$\tau$, $K$ refers to the number of iterations used for PGD. The $\tau$ variable is a hyperparamter used in training which stops the PGD generation of adversarial examples earlier than the normal $K$ number of steps, if the sample is already misclassified.     

\textbf{Why we selected it:} There are many different defenses that rely on adversarial training~\cite{madry2018towards, zhang2019theoretically, wang2019improving, maini2020adversarial} and training and testing them all is not computationally feasible. We selected FAT for its good trade off between clean accuracy and robustness, and because we wanted to test adversarial training on both Vision Transformer and CNN models. In this regard, FAT is one of the adversarial training methods that has already been demonstrated to work across both types of architectures~\cite{mahmood2021robustness}.


\subsection{Trash is Treasure} 

One early direction in adversarial defense design was  model ensembles~\cite{pang2019improving}. However, due to the high transferability of adversarial examples between models, such defenses were shown to not be robust~\cite{tramer2020adaptive}. Trash is Treasure (TiT)~\cite{TiT} is a two model defense that seeks to overcome the transferability issue by training one model $C_{a}(\cdot)$ on the adversarial examples from another model $C_{b}(\cdot)$. At run time both models are used:
\begin{equation}
    y = C_{a}(\psi(x, C_{b}))
\end{equation}
where $\psi$ is an attack done on model $C_{b}$ with input $x$ and $C_{a}$ is the classifier that makes the final class label prediction on the adversarial example generated by $\psi$ with $C_{b}$.

\textbf{Why we selected it:} TiT is one of the newest defenses that tries to achieve robustness in a way that is fundamentally different than pure randomization strategies or direct adversarial training. In our paper, we further develop two versions of TiT. One version is based on the original proposed CNN-CNN implementation. We also test a second mixed architecture version using Big Transfer model and Vision Transformers to try and leverage the low transferability phenomena described in~\cite{mahmood2021robustness}.    

\subsection{Novel Architectures} 

In addition to adversarial machine learning defenses, we also include several novel architectures that have recently achieved state-of-the-art or near state-of-the-art performance in image recognition tasks. These include the Vision Transformer (ViT)~\cite{VIT} and Big Transfer models (BiT)~\cite{BIT}. Both of these types of models utilize pre-training on larger datasets and fine tuning on smaller datasets to achieve high fidelity results. We also test Spiking Neural Network (SNNs) architectures. SNNs are a competitor to artificial neural networks that can be described as a linear time invariant system with a network structure that employs non-differentiable activation functions~\cite{xu2022securing}. A major challenge in SNNs has been matching the depth and model complexity of traditional deep learning models. Two approaches have been used to overcome this challenge, the Spiking Element Wise (SEW) ResNet~\cite{SEW} and transferring weights from existing CNN architectures to SNNs~\cite{rathi2021diet}. We experiment with both approaches in our paper.  

\textbf{Why we selected it:} The set of adversarial examples used to attack one type of architecture (e.g. a ViT) have shown to not be misclassified by other architecture types (e.g. a BiT or SNN)~\cite{mahmood2021robustness, xu2022securing}. While certain white-box attacks have been used to break multiple undefended models, it remains an open question if different architectures combined with different defenses can yield better performance.  


%% file: Sections/Attacks.tex
\section{New Adversarial Attacks}
\label{sec:attack}
In our paper we assume a white-box adversarial threat model. This means the attacker is aware of the set of all defenses $D$ that the defender may use for prediction. In addition, $\forall d \in D$ the attacker also knows the classifier weights $\theta_{d}$, architecture and  any input image transformations the defense may apply. To generate adversarial examples the attacker solves the following optimization problem:
\begin{equation}
\begin{array}{cc}
    \displaystyle \max_{\delta} \sum_{d \in D} L_{d}(x + \delta, y; d) &  \text{subject to: } ||\delta||_p \leq \epsilon
\end{array}
\label{attackform}
\end{equation}
where $D$ is the set of all possible defenses (models) under consideration in the attack, $L_{d}$ is the loss function associated with defense $d\in D$, $\delta$ is the adversarial perturbation, and $(x,y)$ represents the original input with corresponding class label. This is a more general formulation of the optimization problem allowing the attacker to attack single or multi-model classifiers. The magnitude of this perturbation $\delta$ is typically limited by a certain $l_{p}$ norm. In this paper we analyze the attacks and defenses using the $l_{\infty}$ norm. 

Static white-box attacks such as the Projected Gradient Descent (PGD)~\cite{madry2018towards} attack often perform poorly against randomized defenses such as BaRT or TiT. In~\cite{TiT} they tested the TiT defense against an attack designed to compensate for randomness, the Expectation over Transformation attack (EOT) attack~\cite{EOT}. However, it was shown that the EOT attack performs poorly against TiT (e.g. $20\%$ or worse attack success rate). For attacking BaRT, in~\cite{demystifying} they proposed a new white-box attack to break BaRT. However, this new attack requires that the image transformations used in the  BaRT defense be differentiable, which is a deviation from the original BaRT implementation. 

\textbf{Attack Contributions:} It is crucial for both the attacker and defender to consider the strongest possible adversary when playing the adversarial examples game. Thus, we propose three new white-box attacks for targeting randomized defenses. The first attack is designed to work on single, randomized defenses and is called the Momentum Iterative Method over Expectation (MIME). To the best of our knowledge, MIME is the first white-box attack to achieve a high attack success rate ($>70\%$) against TiT. MIME is also capable of achieving a high attack success rate against BaRT, even when non-differentiable transformations are implemented as part of the defense. Our second attack, is designed to generate adversarial examples that work against multiple type of defenses simultaneously. This compositional attack is called, the Multi-Agent Gradient Expectation attack (MAGE). Lastly, we propose a modified version of MAGE called MAGE-D which is designed to target defenses utilizing consensus voting adversarial detection techniques. MAGE-D is discussed in Section~\ref{sec:GaME}.

\subsection{Momentum Iterative Method over Expectation} 
We develop a new white-box attack specifically designed to work on defenses that inherently rely on randomization, like Barrage of Random Transforms (BaRT)~\cite{BaRT} and Trash is Treasure (TiT)~\cite{TiT}. Our new attack is called the Momentum Iterative Method over Expectation (MIME). The attack ``mimes" the transformations of the defender in order to more precisely model the gradient of the loss function with respect to the input after the transformations are applied. To this end, MIME utilizes two effective aspects from earlier white-box attacks: momentum from the Momentum Iterative Method (MIM)~\cite{MIM} attack, and repeated sampling~\cite{EOT} from the Expectation Over Transformation (EOT) attack:
\begin{equation}
    x^{(i)}_{adv} = x^{(i-1)}_{adv} + \epsilon_{\text{step}} g^{(i)}
    \label{eq:mime}
\end{equation}
where the attack is computed iteratively with $x^{(0)}_{adv}=x$. In Equation~\ref{eq:mime} $g^{(i)}$ is the momentum based gradient of the loss function with respect to the input at iteration $i$ and is defined as:
\begin{equation}
    g^{(i)} := \gamma g^{(i-1)} + \mathbb{E}_{t \sim T}[\frac{\partial L}{\partial t(x^{(i)}_{adv})}]
    \label{eq:MIMEMomentum}
\end{equation}
where $\gamma$ is the momentum decay factor and $t$ is a random transformation function drawn from the defense's transformation distribution $T$. In Table~\ref{tbl:MIME}, we show experimental results for the MIME attack on CIFAR-10 randomized defenses (TiT and BaRT). It can clearly be seen that MIME has a higher attack success rate than both APGD~\cite{APGD} and MIM~\cite{MIM}.

\begin{table}[!h]
\centering

\resizebox{.45\textwidth}{!}{
\begin{tabular}{|c|ccccc|}
\hline
Attack  & BaRT-1          & BaRT-5          & BaRT-10         & TiT (BiT/ViT)  & TiT (VGG/RN) \\ \hline
MIME-10 & \textbf{3.18\%} & 15.5\%          & 43.2\%          & 10.1\%         & 24.9\%            \\
MIME-50 & 4.3\%           & \textbf{8.22\%} & \textbf{23.2\%} & \textbf{8.3\%} & \textbf{23.3\%}   \\
MIM     & 6.7\%           & 39.5\%          & 59.5\%          & 52\%           & 58.9\%            \\
APGD    & 8.9\%           & 47.7\%          & 70.8\%          & 68.2\%         & 40.7\%            \\
Clean   & 98.4\%          & 95.3\%          & 92.5\%          & 90.1\%         & 76.6\%            \\ \hline
\end{tabular}}
\caption{Performance of the MIME attack against CIFAR-10 randomized defenses TiT and BaRT with $\epsilon_{max}=0.031$. It can clearly be seen that MIME outperforms both APGD and MIM on these two randomized defenses. Full experimental details are given in the supplemental material.}
\label{tbl:MIME}
\end{table}

\subsection{Multi-Agent Gradient Expectation attack}


The use of multi-model attacks are necessary to achieve a high attack success rate when dealing with ensembles that contain both CNN and non-CNN model architectures, like the Vision Transformer (ViT)~\cite{VIT} and Spiking Neural Network (SNN)~\cite{SEW}. This is because adversarial examples generated by single model white-box attacks generally do not transfer well between CNNs, ViTs and SNNs~\cite{mahmood2021robustness, xu2022securing}. In addition it is an open question if multi-model attacks can be effective against the current state-of-the-art defenses. In this paper, we expand the idea of a multi-model attack to include not only different architecture types, but also different defenses. The generalized form of the multi-model attack is found in Equation~\ref{attackform}.

For a single input $x$ with corresponding class label $y$, an untargeted multi-model attack is considered successful if $(\forall d \in D, C_{d}(x+\delta) \neq y) \land (||\delta||_p \leq \epsilon)$. One formulation of the multi-model attack is the Auto Self-Attention Gradient Attack (Auto SAGA)~\cite{xu2022securing}. Auto SAGA iteratively attacks combinations of ViTs, SNNs and CNNs:
\begin{equation}
    x_{adv}^{(i+1)} = x_{adv}^{(i)} + \epsilon_{step}*\text{sign}(
    G_{blend}(x_{adv}^{(i)}))
    \label{eq:sagamain}
\end{equation}
where $\epsilon_{step}$ was the step size used in the attack. In the original formulation of Auto SAGA, $G_{blend}$ was a weighted average of the gradients of each model $d \in D$. By combining gradient estimates from different models, Auto SAGA is able to create adversarial examples that are simultaneously misclassified by multiple models. One limitation of Auto SAGA attack is that it does not account for defenses that utilize random transformations. Motivated by this, we can integrate the previously proposed MIME attack into the gradient calculations for Auto SAGA. We denote this new attack as the Multi-Agent Gradient Expectation attack (MAGE). Both SAGA and MAGE use the same iterative update (Equation~\ref{eq:sagamain}). However, MAGE uses the following gradient estimator:
\begin{equation}
    {\small
    \begin{array}{c}
         \displaystyle G_{blend}(x_{adv}^{(i)}) = \gamma G_{blend}(x_{adv}^{(i-1)}) + \sum_{k \in D \backslash R} \alpha^{(i)}_{k} \phi^{(i)}_{k} \odot \frac{\partial L_{k}}{\partial x_{adv}^{(i)}}  \\
         \displaystyle + \sum_{r \in R} \alpha^{(i)}_{r} \phi^{(i)}_{r} \odot (\mathbb{E}_{t \sim T}[\frac{\partial L_{r}}{\partial t(x^{(i)}_{adv})}])
    \end{array}}
    \label{eq:ae}
\end{equation}
In Equation~\ref{eq:ae}, the two summations represent the gradient contributions of sets $D \backslash R$ and $R$, respectively. Here we define $R$ as the set of randomized defenses and $D$ as the set of all the defenses being targeted. In each summation $\phi$ is the self-attention map~\cite{abnar2020quantifying} which is replaced with a matrix of all ones for any defense that does not use ViT models. $\alpha_{k}$ and $\alpha_{r}$ are the associated weighting factors for the gradients for each deterministic defense $k$ and randomized defense $r$, respectively. Details of how the weighting factors are derived are given in~\cite{xu2022securing}. We further develop this  attack formulation into a novel attack (MAGE-D) that is effective on detection based defenses in Section~\ref{sec:GaME}. Experimental results for both MAGE and MAGE-D are give within Section~\ref{sec:exp}.

%% file: Sections/ExperimentsTransfer.tex
\section{Transferability Experiments}
\label{sec:trans}
\begin{table}
    \centering
    \resizebox{.8\linewidth}{!}{
    \begin{tabular}{|c|ccc|}
            \hline
            Defense & Best Attack & Clean Acc & Robust Acc \\ \hline
            B1      & MIME        & 98.40\%   & 3.40\%     \\ \hline
            B5      & MIME        & 95.30\%   & 15.00\%    \\ \hline
            B10     & MIME        & 92.50\%   & 43.50\%    \\ \hline
            RF      & APGD        & 81.89\%   & 52.00\%    \\ \hline
            VF      & APGD        & 92.36\%   & 25.00\%    \\ \hline
            ST      & APGD        & 91.54\%   & 0.00\%     \\ \hline
            SB      & APGD        & 81.16\%   & 1.60\%     \\ \hline
            BVT     & MIME        & 90.10\%   & 8.60\%     \\ \hline
            VRT     & MIME        & 76.60\%   & 26.20\%    \\ \hline
        \end{tabular}}
        \caption{Single defense implementations for CIFAR-10 with the corresponding strongest attack on the defense and the clean accuracy of the defense. All attacks are done with $\epsilon_{max}=0.031$. Complete attack and defense implementation details are given in our supplementary material.}
        \label{tbl:bestAttack}
\end{table}
\begin{figure}
    \centering
    \includegraphics[width=60mm]{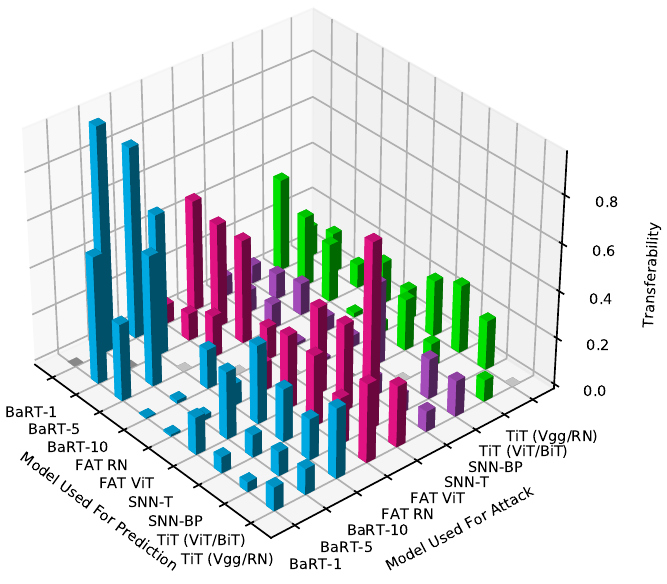}
    \captionsetup{type=figure}
    \caption[]{Visual representation of the transferability of adversarial examples between defenses for CIFAR-10. A higher bars means more adversarial examples are misclassified. Full numerical tables for the figure are given in our supplemental material.}
    \label{fig:transferability}
\end{figure}
Adversarial transferability refers to the phenomena in which adversarial examples generated to attack one model are also misclassified by a different model. Adversarial transferability studies have been done on a variety of machine learning models~\cite{Delving, mahmood2021robustness, xu2022securing}. However, to the best of our knowledge, adversarial transferability between different state-of-the-art defenses has not been conducted. This transferability property is of significant interest because a lack of transferability between different defenses may indicate a new way to improve adversarial robustnes. 

In Table~\ref{tbl:bestAttack}, we show the different single defenses we analyze in this paper and the best single attack on each of them from the set of attacks (MIM~\cite{MIM}, APGD~\cite{APGD} and MIME (proposed in this work). In Figure~\ref{fig:transferability}, we visually show the transferability results of these attacks for CIFAR-10. The blue bars represent the transferability of adversarial examples generated using the best attack for BaRT models, the green bars represent adversarial examples generated using  the best attack for TiT, the pink bars represent the adversarial examples generated using the best attack for FAT and lastly the purple bars represent the adversarial examples generated using SNNs. We give detailed discussions of these results in our supplementary material and briefly summarize of the key takeaways from these experiments.

Adversarial examples generated using the best attack on one defense \textit{did not} transfer well to other defenses. For example, only $0.8\%$ of the adversarial examples generated by the BaRT-1 defense transfer to the FAT ViT defense. The average transferability for the 8 different defenses shown in Figure~\ref{fig:transferability} is only $21.62\%$ and there is no single model attack that achieves strong performance (i.e. $>50\%$) across all defenses. These results in turn motivate the development of a game theoretic framework for both the attacker and defender. For the attacker, this prompts the need to use multi-model attack like MAGE that was proposed in Section~\ref{sec:attack}, as no single attack (APGD, MIM or MIME) is ubiquitous. For the defender, these results highlight the opportunity to increase robustness by taking advantage of the low levels of transferability between defenses through the implementation of a randomized ensemble defense.

%% file: Sections/Game.tex
\section{Game theoretic Mixed Experts (GaME)}
\label{sec:GaME}
In this section we derive our framework, Game theoretic Mixed Experts (GaME), for approximating a Nash equilibrium in the adversarial examples game. In comparison to other works \cite{meunier2021mixed} \cite{pinot2020randomization} \cite{pal2020game} \cite{balcan2022nash} \cite{pmlr-v151-le22c} we take a more discretized approach and solve the approximate version of the adversarial examples game. This ultimately leads to the creation of a finite, tabular, zero-sum game that can be solved in polynomial time using linear programming techniques.
\subsection{The Adversarial Examples Game}
We build upon and discretize the adversarial examples game explored in~\cite{meunier2021mixed}. The adversarial examples game is a zero-sum game played between two players: the attacker, $p_A$, and the defender $p_D$. Let $\mathcal{X}$ be the input space and $\mathcal{Y}$ the output space of $p_D$'s classifiers, and let $\Theta$ represent the space of classifier parameters. Additionally, let $P_{\epsilon, \mathcal{X}} = \{x \in \mathcal{X} : ||x||_p \leq \epsilon \}$ be the set of valid adversarial perturbations for norm $p$ and $\epsilon \in \mathbb{R}^+$.
Let $A^*_\epsilon = \{(f : \Theta \times \mathcal{X} \times \mathcal{Y} \rightarrow P_{\epsilon, \mathcal{X}})\}$ be the set of all valid attack functions. The goal of $p_A$ is to choose $a \in A^*_\epsilon$, which maximizes the expected loss of $p_D$'s classifier, $\theta$, given some pair of input and ground truth label $(x,y) \in \mathcal{X} \times \mathcal{Y}$. The goal of $p_D$ is to minimize this loss through its choice of $\theta \in \Theta$. We can thus formulate the adversarial examples game as a mini-max optimization problem:
\begin{equation}
    \displaystyle \inf_{\theta \in \Theta} \sup_{a \in A^*_\epsilon} \E_{(x,y) \sim \mathcal{X} \times \mathcal{Y}} [L(x + a(\theta ,x,y), y ; \theta)]
\end{equation}
Due to the vastness of $\Theta$ and $A^*_\epsilon$, solving this optimization problem directly is currently computationally intractable. To this end, in the next subsections we will formulate $\text{GaME}_1$ and $\text{GaME}_n$ which discretize $\Theta$ and $A^*_\epsilon$ by enlisting a set of state-of-the-art attacks and defenses. 
\subsection{$\textbf{GaME}_1$}
The goal of the GaME framework is to find an approximate solution to the adversarial examples game through the implementation of a set of attacks and defenses which will serve as experts for $p_A$ and $p_D$. Let $A'  \subset  A^*_\epsilon$ be a subset of all valid adversarial attack functions chosen by $p_A$. Additionally, let $D \subset \Theta$ be a set of defense classifiers chosen by $p_D$. We further impose that all $a\in A'$ are white-box attacks (see Section~\ref{sec:attack} for our adversarial threat model) and that $A',D$ are finite, i.e. $|A'| \leq N_a$ and $|D| \leq N_d$ for some $N_a, N_d \in \mathbb{N}$. It is important to note that each $a \in A'$ is a function of some classifier, $\theta \in \Theta$, in addition to the input and ground truth label. Due to this it is possible for $p_A$ to chose to attack defense $d \in D$ with attack $a \in A'$, while $p_D$ chooses to evaluate the sample using defense $d' \in D$ where $d \neq d'$. Therefore, for convenience, we will define a new, more general set of attack strategies for $p_A$:
\begin{equation}
    \begin{array}{c}
         A \subseteq \{ (f : \mathcal{X} \times \mathcal{Y} \rightarrow P_{\epsilon, \mathcal{X}}) : \\ 
         f(x,y) = a_i(U, x, y), \; a_i \in A', \; U \subseteq D \}  
    \end{array}
\end{equation}
where we extend the definition of $A'\subseteq A^*_\epsilon$ to attack functions, such as MAGE, that can take subset of defense parameters $U\subseteq D$ as input (see Equation~\ref{attackform} for our multi-model attack formulation). Thus we will let $D$ be the strategy set of $p_D$, and $A$ be the strategy set of $p_A$. We can then formulate a discretized version of the adversarial examples game as follows:
\begin{equation}
    \displaystyle \min_{d \in D} \max_{a \in A} \E_{(x,y) \sim \mathcal{X} \times \mathcal{Y}}[L(x + a(x,y), y ; d)]
\end{equation}
In the above two formulations we optimize over the set of pure strategies for the attacker and defender. However, as previously explored in~\cite{araujo2020advocating}~\cite{meunier2021mixed}, limiting ourselves to pure strategies severely inhibits the strength of both the attacker and defender. Thus we create the following mixed strategy vectors for $p_A$, $p_D$:
\begin{equation}
\begin{array}{c}
    \lambda^A \in \{r \in [0,1]^{|A|} : ||r||_1 = 1\} \\    \lambda^D \in \{r \in [0,1]^{|D|} : ||r||_1 = 1\}
\end{array}
\end{equation}
here $\lambda^A$ and $\lambda^D$ represent the mixed strategies of $p_A$ and $p_D$ respectively. Let each $a_i \in A$ and $d_i \in D$ correspond to the $i^{th}$ elements of $\lambda^A$ and $\lambda^D$, $\lambda^A_i$ and $\lambda^D_i$, respectively.
\begin{equation}
    \mathbb{P}(\{a_i \in A : a = a_i\}) = \lambda^A_i, \;  
    \mathbb{P}(\{d_i \in D : d = d_i\}) = \lambda^D_i
\end{equation}
where $a \in A$ and $d \in D$ are random variables. With these mixed strategy vectors we can then reformulate the adversarial examples game as a mini-max optimization problem over $p_D$'s choice of $\lambda^D$ and $p_A$'s choice of $\lambda^A$:
\begin{equation}
{\small
\begin{array}{c}
    \displaystyle \min_{\lambda^D} \max_{\lambda^A} \E_{(x,y) \sim \mathcal{X} \times \mathcal{Y}}[ \E_{(a,d) \sim A \times D} [L(x + a(x,y), y ; d)]] =  \\
    \displaystyle  \min_{\lambda^D} \max_{\lambda^A} \E_{(x,y) \sim \mathcal{X} \times \mathcal{Y}}[ \sum_{a_i \in A} \lambda^A_i 
    \displaystyle \sum_{d_i \in D} \lambda^D_i [L(x + a_i(x,y), y ; d_i)]]
\end{array}}
\end{equation}
For continuous and or non-finite $\mathcal{X}$, $D$, and $A$ solving the above optimization problem is currently computationally intractable. Thus, we can instead approximate the mini-max optimization by taking $N$ Monte-Carlo samples with respect to $(x_j,y_j) \in \mathcal{X} \times \mathcal{Y}$:
{\small \begin{equation}
   \displaystyle  \min_{\lambda^D} \max_{\lambda^A} \frac{1}{N}\sum_{j=0}^N \sum_{a_i \in A} \lambda^A_i \sum_{d_i \in D} \lambda^D_i [L(x_j + a_i(x_j,y_j), y_j ; d_i)]
\end{equation}}
For convenience we denote $r_{d_i, a_i} = \frac{1}{N}\sum_{j=0}^N [L(x_j + a_i(x_j,y_j), y_j ; d_i)]$. Colloquially $r_{d_i, a_i}$ represents the expected robustness of defense $d_i$ when evaluating adversarial samples generated by attack $a_i$. From a game theoretic perspective, $r_{d,a}$ is the payoff for $p_D$ when they play strategy $d$ and $p_A$ plays strategy $a$. The payoff for $p_A$ given strategies $d,a$ is $-r_{d,a}$. Our mini-max optimization problem can then be simplified to:
\begin{equation}
   \displaystyle  \min_{\lambda^D} \max_{\lambda^A} \sum_{a_i \in A} \lambda^A_i \sum_{d_i \in D} \lambda^D_i r_{d_i, a_i}
\end{equation}
From this we can create a finite, tabular, zero-sum game defined by the following game-frame in strategic form:
\begin{equation}
    \langle \{p_A, p_D\}, \; (A,D), \; O, \; f \rangle
\end{equation}
where $O = \{ r_{d,a} \; \forall a \in A, \; d \in D \}$ and $f$ is a function $f : A \times D \rightarrow O$ defined by $f(d,a) = r_{d,a}$. Because this is a finite, tabular, zero-sum game, it has a Nash-Equilibrium~\cite{Nash}. Let $R$ be the payoff matrix for $p_D$ where $R_{d,a} = r_{d,a}$. It then becomes the goal of $p_D$ to maximize their guaranteed, expected payoff. Formally, $p_D$ must solve the following optimization problem:
\begin{equation}
\begin{array}{c c}
    \displaystyle \max_{r^* ; \lambda^D} r^* \,  & \text{subject to } \lambda^D R \geq (r^*, \cdots r^*) \text{ and } ||\lambda^D||_1 \leq 1
\end{array}
\end{equation}
This optimization problem is a linear program, the explicit form of which we provide in the supplemental material. All linear programs have a dual problem, in this case the dual problem finds a mixed Nash strategy for $p_A$. This can be done by changing the problem to a minimization problem and transposing $R$. In the interest of space we give the explicit form of the dual problem in the supplemental material as well. These linear programs can be solved using polynomial time algorithms. 
\subsection{$\textbf{GaME}_n$}
$\text{GaME}_n$ is a more general family of games of which $\text{GaME}_1$ is a special case. In $\text{GaME}_n$, for $n>1$, $p_D$ can calculate their final prediction based upon the output logits of multiple $d \in D$ evaluated on the same input $x$. For this to occur, $p_D$ must also choose a function to map the output of multiple defenses to a final prediction. Formally, the strategy set of $p_D$ becomes $D = D' \times F$, where $F$ is a set of prediction functions and $D'$ is defined as follows.
\begin{equation}
    D' \subseteq \{U : U \subseteq D, \; |U| \leq n\}
\end{equation}
Multi-model prediction functions can increase the overall robustness of an ensemble by requiring an adversarial sample to be misclassified by multiple models simultaneously~\cite{BARZ}. In this paper we focus on two voting functions: the majority vote ($f^h$)~\cite{BaRT}, and the largest softmax probability vote ($f^s$)~\cite{demystifying}:
\begin{equation}
    f^h(x, U) = \displaystyle \argmax_{y \in \mathcal{Y}} \sum_{d \in U} \mathbbm{1} \{ y = \argmax_{j \in \mathcal{Y}} d_j(x)\}
    \label{eq:fh}
\end{equation}
\begin{equation}
    f^s(x, U) = \displaystyle \argmax_{y \in \mathcal{Y}} \frac{1}{|U|} \sum_{d \in U} \sigma(d(x))
    \label{eq:fs}
\end{equation}
where $\mathbbm{1}$ is the indicator function, $\sigma$ is the softmax function, and $d_j(x)$ represents the $j^{th}$ output logit of defense $d$ when evaluated on input $x$. Solving $\text{GaME}_n$ is the same as solving $\text{GaME}_1$, except with larger $|D|$. Notationally, the mini-max optimization problem in terms of $r_{d,a}$ remains the same as we have simply redefined $D$, however we can redefine $r_{d,a}$ as follows:
\begin{equation}
    r_{(U_i, f_j), a_k} = \sum_{l=0}^N [L(x_l + a_k(x_l,y_l), y_l ; (U_i, f_j))]
\end{equation}
where $(U_i, f_j) \in D = D' \times F$. Similarly to $\text{GaME}_1$, in $\text{GaME}_n$  $r_{(U_i,f_j),a_k}$ represents the expected robust accuracy, i.e., the payoff, for $p_D$ if they play strategy $(U_i,f_j)$ and $p_A$ plays strategy $a_k$.
\subsection{$\textbf{GaME-D}_n$ and MAGE-D}
Motivated by the low transferability of adversarial examples, as shown in Section~\ref{sec:trans}, we enhance the GaME framework to include adversarial detection. We denote this expanded framework as $\text{GaME-D}_n$. In the expanded framework, the defender $p_{d}$ is allowed to output an additionally adversarial label, $\bot$. The condition for a successful adversarial attack conducted by $p_{a}$ is then: $( f^{d}(x_{adv},D) \neq y) \land ( f^{d}(x_{adv},D)\neq \bot)$ where $D$ is the set of defenses and $(x,y)$ is the clean input and correct class label used to generate $x_{adv}$. In our setup we empirical found plurality voting to provide the best trade-off between robustness and clean accuracy. The plurality voting utilized by $p_{d}$ is: 
\begin{equation}
\begin{array}{c}
f^d(x, U) = \left\{ \begin{array}{c c}
    y'_i & \text{if } \exists \; V_{y'_i} : \forall y'_j \neq y'_i, \; V_{y'_i} > V_{y'_j} \\ 
    \bot & \text{otherwise}
\end{array}    \right. \\
\end{array}
\end{equation}
where $V_{y'_i}$ represents the total number of votes for class label $y'_i$ and $\bot$ represents the adversarial class label. Under $\text{GaME-D}_n$, the defender optimizes $r_{(U_i, f_j), a_k}$ for both clean accuracy and robustness (e.g. without optimizing over clean accuracy $p_{d}$ can always output $\bot$, achieving $100\%$ robustness but $0\%$ clean accuracy). Therefore, for both $\text{GaME}_n$ and $\text{GaME-D}_n$, we adjust the set of attacks $A$ such that $f^{identity} \in A$ where $f^{identity} : \mathcal{X} \times \mathcal{Y} \rightarrow P_{\epsilon, \mathcal{X}}$ defined by $ f^{identity}(x,y) = x$. 
\begin{algorithm}
\caption{MAGE} \label{alg:MAGE}
\begin{algorithmic}
\State {\textbf{Input}: clean sample $x$, true label $y$, boolean $targeted$.} 
\State \textbf{Initialize}: $G_{blend} = \hat{0}$ 
\State \textbf{Define}: $g(m, x) = \sum_{j=0}^{N_{samp}} \frac{\partial L_{m}}{\partial t_m(x_{adv}^{(i)})}$
\State \textbf{For} $i$ in range 1 to $N_{iter}$ do:
\State \hspace{10pt} $G_{blend} = \gamma \cdot G_{blend} + (\sum_{m=1}^{M}\alpha_{m}^{i}\odot g(m, x_{adv}^{(i)}) \odot \phi_{m})$
\State \hspace{10pt} \textbf{If} $targeted$
\State \hspace{20pt} $x_{adv}^{(i+1)}=x_{adv}^{(i)}-\epsilon_{step}\text{sign}(G_{blend})$
\State \hspace{10pt} \textbf{else}
\State \hspace{20pt} $x_{adv}^{(i+1)}=x_{adv}^{(i)}+\epsilon_{step}\text{sign}(G_{blend})$
\State \hspace{10pt} $x_{adv}^{i+1}=p(x_{adv}^{(i)}, x, \epsilon_{max})$
\State \hspace{10pt} \textbf{For} $m$ in range 1 to M:
\State \hspace{20pt}${\frac{\partial x_{adv}^{(i)}}{\partial \alpha _{m}^{(i)}}} \approx u\epsilon _{step} \text{sech}^{2}(u\sum_{m=1}^{M}  g(m, x_{adv}^{(i)}) \odot $
\State \hspace{50pt} $g(m, x_{adv}^{(i)}))$
\State \hspace{20pt}${\frac{\partial F}{\partial \alpha _{m}^{(i)}}}=
{\frac{\partial F}{\partial \alpha _{adv}^{(i)}}}\odot 
\frac{\partial x_{adv}^{(i)}}{\partial \alpha _{m}^{(i)}}$
\State \hspace{20pt}$\alpha _{m}^{(i)}=\alpha _{m}^{(i)}-r{\frac{\partial F}{\partial \alpha _{m}^{(i)}}}$
\State \hspace{10pt} end for
\State  end for
\State {\textbf{Output}:} $x_{adv}$
\end{algorithmic}
\end{algorithm} 
\begin{algorithm}
\caption{MAGE-D}\label{alg:MAGE-D}
\begin{algorithmic}
\State {\textbf{Input}: clean sample $x$, true label $y$.} 
\State \textbf{Let} $\sigma (m,x)$ be a function that gets the output logits of a classifier $m$ given input $x$
\State \textbf{Initialize} $y_{adv}$, $x'_{adv}$, and $s$ to be lists indexed by $m \in M$
\State $x_{adv}$ = MAGE($x$, $y$, False)
\State \textbf{If} $D(x_{adv}) \neq y$ and $D(x_{adv}) \neq null$
    \State \hspace{10pt} \textbf{return} $x_{adv}$
\State \textbf{For} $m \in M$
    \State \hspace{10pt} $y'_{m} = m(x_{adv})$
    \State \hspace{10pt} \textbf{If} $y' \neq y$
        \State \hspace{20pt} $x'_{adv, m} =$ MAGE($x$, $y'_m$, True)
        \State \hspace{20pt} $s_m = \sigma(D, x'_{adv, m})$
\State  $m^* = \argmax_{m \in M} [\max_{y'} s_{m, y' \neq y}]$
\State \textbf{return} $x'_{adv,m^*}$
\end{algorithmic}
\end{algorithm}
MAGE, MIME, and APGD, are agnostic to the adversarial detection defenses. To represent a stronger adversary, we propose a modification to the MAGE attack, MAGE-D. The attack exploits a key property of the detector's voting scheme: by leveraging a targeted version of MAGE, the attacker can work to achieve a plurality of votes for the wrong class label. A detailed description of the algorithm can be found in Algorithms~\ref{alg:MAGE} and~\ref{alg:MAGE-D}. For this attack, we define the attack parameters as follows: number of iterations $N_{iter}$, number of samples $N_{samp}$, step size per iteration $\epsilon_{iter}$, maximum perturbation  $\epsilon_{max}$, coefficient learning rate $r$, and detector, $D$, utilizing a set of models, $M$, with corresponding loss functions $L_{m_1},..,L_{m_{|M|}}$ and transform functions $t_{m_1},...,t_{m_{|M|}}$. Additionally we utilize a projection function $p: \mathcal{X} \rightarrow P_{\epsilon_{max}, \mathcal{X}}$ which projects an adversarial example, $x_{adv}$, generated from clean sample, $x$, such that $|x_{adv} - x|_\infty \leq \epsilon_{max}$.

MAGE-D is able to increase the attack success rate of MAGE by 50\% on CIFAR-10 and 85.3\% on Tiny ImageNet when attacking a detector utilizing vanilla voting models. Further details for these results can be found in the supplementary material.

%% file: Sections/ExperimentsGame.tex
\section{Experimental Results} \label{sec:exp}
\begin{figure}
\centering
\subfloat[CIFAR-10]{%
  \includegraphics[clip,width=.9\columnwidth]{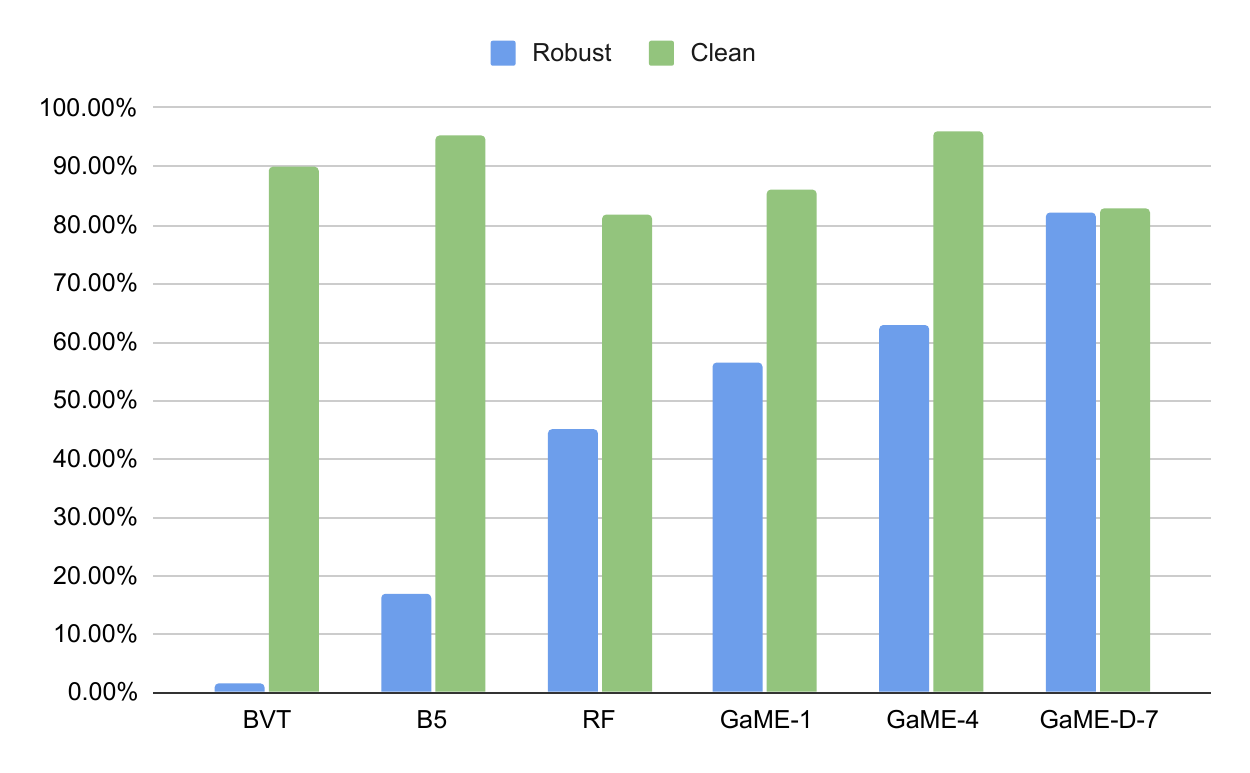}%
}

\subfloat[Tiny ImageNet]{%
  \includegraphics[clip,width=0.9\columnwidth]{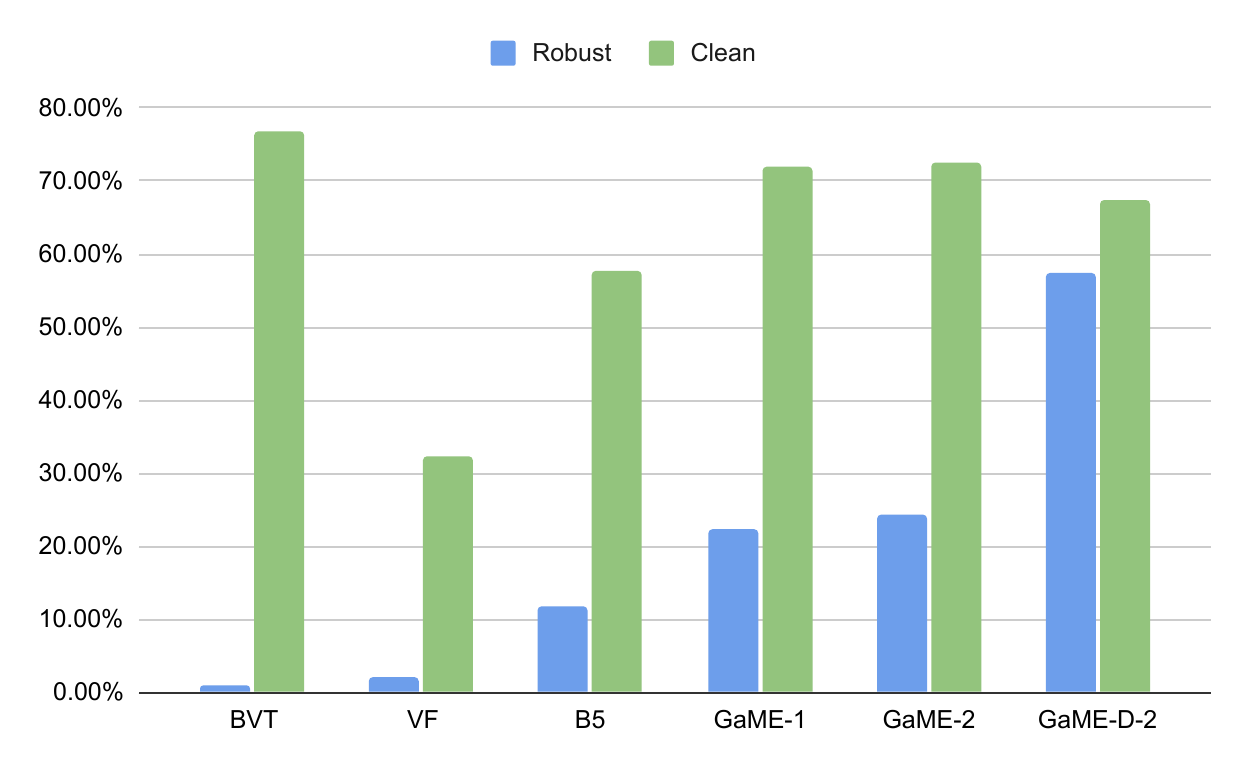}%
}
\caption{Comparison of GaME ensembles to the three most robust, single-model defenses for each dataset. and GaME ensemble results.}
\label{bars}
\end{figure}
 We test on two datasets for our experimental results, CIFAR-10~\cite{c10} and Tiny ImageNet~\cite{le2015tiny}. For CIFAR-10 we solved instances of $\text{GaME}_n$ using the following defenses: BaRT-1 (B1), BaRT-5 (B5), ResNet-164-FAT (RF), ViT-L-16-FAT (VF), SNN Transfer (ST), Backprop SNN (SB), and TiT using ViT and BiT (BVT). For Tiny ImageNet we solved instances of $\text{GaME}_n$ utilizing: BaRT-1, BaRT-5, ViT-L-16-FAT, and TiT using ViT and BiT. Explicit details for our experimental setup are given in the supplementary material.
 
In Figure~\ref{bars} and throughout the paper, we measure the robustness of each ensemble as the lowest robust accuracy achieved by any single attack in our study. From a game-theoretic perspective this can be seen as the minimum, guaranteed utility for the defender. Of all the defenses in our study the FAT ResNet-164 had the highest robust accuracy on CIFAR-10 at 45\%. On Tiny ImageNet, BaRT-5 had the highest robust accuracy at 11\%. Compared to these defenses our GaME generated ensemble achieved 63\% robustness on CIFAR-10 and 24.5\% robustness on Tiny ImageNet without using detectors. When detectors were implemented this increased to to 82.31\% robustness and 57.40\% robustness on each respective dataset.

Additionally, our ensemble is able to maintain a high level of clean accuracy. In particular, the GaME framework maintains 96.2\% clean accuracy on CIFAR-10 and 72.6\% clean accuracy on Tiny ImageNet without detectors. This is only out performed by BaRT-1 with a clean accuracy of 98.4\% on CIFAR-10 and the BiT-ViT Trash is Treasure defense with a clean accuracy of 76.97\% on Tiny ImageNet. With detectors the clean accuracy dropped to 82.9\% and 67.3\% respectively. It is important to note that in both these cases the robustness of the GaME framework is much higher both with and without adversarial detection. We provide more detailed, numerical results for these figures in the supplementary material. Additionally, we provide studies of the effects of $n$ (the maximum ensemble size) and $N$ (the sample number) on $\text{GaME}_n$ generated ensembles along with an analysis of the computational cost of the framework. Lastly, we include studies comparing the effectiveness of the GaME framework to naive ensemble defenses utilizing a uniform distribution over all strategies. 


%% file: Sections/Conclusion.tex
\section{Conclusion}
The field of adversarial machine learning has begun to cycle between new defenses, specialized attacks to break those defenses, and further specialized defenses to mitigate the attacks. In this paper, we seek to go beyond this cat and mouse dynamic by considering a new game theoretic framework that incorporates multiple defenses, adversarial detection, and compositional attacks.

We develop three new white-box attacks, the Momentum Iterative Method over Expectation (MIME) for attacking single randomized defenses and the Multi-Agent Gradient Expectation attacks (MAGE and MAGE-D) for dealing with a combination of randomized and non-randomized defenses and detectors. Additionally, we are the first to show the transferability of adversarial examples generated by MIM, APGD, MIME and MAGE on state-of-the-art defenses like FAT, BaRT and TiT. 

Lastly, and most importantly, we develop a game theoretic framework to approximate an optimal strategy for adversarial attackers and defenders. The flexibility of this framework allows any newly proposed defense or attack to be easily integrated. Using a set of SOTA attacks and defenses we demonstrate that our game theoretic framework can create a compositional defense that achieves a 37.3\% and 45.65\% increase in robustness on Tiny ImageNet and CIFAR-10 respectively when utilizing a mixed Nash strategy (compared to using the best single defense). Both compostional GaME defenses also come with higher a clean accuracy than the most robust, single model defenses.

%% file: Sections/Appendix.tex
\section{Supplementary Material}

In the supplementary material we provide additional studies and results that further build on the results presented in the main body of the paper. First, in subsection~\ref{sec:setup} we provide a more explicit attack setup for the experiments, we performed throughout the paper. 

In the next two subsections we perform further experiments on the GaME framework not seen in the main body of the paper. In particular, in subsection~\ref{sec:compS} we provide a study on the computational cost of the $\text{GaME}_n$ framework; and in subsection~\ref{sec:sampleS} we study how the number of samples, $N$, used to approximate the game matrix for $\text{GaME}_n$ effects the ensemble defense robustness.

Following this, the next two subsections further explore the effectiveness of the MAGE and MAGE-D attacks. Namely, in subsection~\ref{sec:MAGE} we provide experimental results showing the effectiveness of MAGE-D over MAGE, when attacking a defender implementing detectors; and in subsection~\ref{sec:threeS} we provide a brief study upon the effectiveness of MAGE, when attacking 3 defenses simultaneously.

In the next two subsections we provide additional implementation details about the MIME attack and GaME framework. Specifically, in subsection~\ref{sec:lpS} we provide explicit forms for the linear programs one must solve to create a $\text{GaME}_n$ defense or attack ensemble; and In subsection~\ref{sec:approxS} we provide an approximate form for the update equation of MIME seen in the main body of the paper since a direct computation of the update is not feasible.

In the last two subsections we give explicit numerical results for the transferability and GaME experiments explored in the main body of the paper. In particular, in subsection~\ref{sec:transS} we provide numerical results and further analysis for the transferability experiments found in the main body of the paper. Lastly, in subsection~\ref{sec:resultsS} we provide further studies and analyses with the GaME framework, including comparisons to uniform probability distribution ensemble defenses, and a study on the effect of $n$ on a $\text{GaME}_n$ defense.

\subsection{GaME Experimental Setup}
\label{sec:setup}
For each version of $\text{GaME}_n$ played on CIFAR-10 we chose to employ the following defenses: BaRT-1, BaRT-5, TiT using BiT and ViT, FAT ResNet-164, FAT ViT-L-16, SEW SNN, and Transfer SNN. BaRT-1 and BaRT-5 were chosen in favor of BaRT-10 due to the computational cost of computing all the necessary attacks on BaRT-10. Additionally, we chose the Bit+ViT version of TiT since the original, VGG-ResNet, architecture that was proposed has a significantly lower clean accuracy~\cite{TiT}. For Tiny ImageNet we chose BaRT-1, BaRT-5, TiT using BiT and ViT, and FAT ViT-L-16.

We attacked every random transform defense with MIME, every non-random-transform defense with APGD, and each pair of defenses with MAGE. This came to a total of 28 attacks on CIFAR-10 and 10 attacks on Tiny ImageNet. Every attack was run with respect to the $l_\infty$ norm. The hyperparameters for each attack are seen in Table~\ref{attackSetup}:
\begin{table}[H]
\centering
\resizebox{.45\textwidth}{!}{\begin{tabular}{|c|ccccccc|}
\hline
Attack  & $\epsilon$ & $\epsilon_{step}$ & Attack Steps & $N$ & $\gamma$ & Fitting Factor & $\alpha$ Learning Rate \\ \hline
APGD    & .031    & .005         & 20           & -           & -              & -              & -                   \\
MIM     & .031    & .0031        & 10           & -           & .5             & -              & -                   \\
MIME    & .031    & .0031        & 10           & 10          & .5             & -              & -                   \\
MAGE & .031    & .005         & 40           & 4           & .5             & 50             & 10000               \\
MAGE-D & .031    & .005         & 40           & 4           & .5             & 50             & 10000               \\\hline
\end{tabular}}
\caption{Attack parameters for each of the attacks used in the paper. Here $\gamma$ represents the momentum decay rate, $\epsilon$ represents the maximum allowed perturbation magnitude, $\alpha$ represents the weights used in the MAGE algorithm, and $N$ represents the number of EOT samples taken. Note for APGD that the $\epsilon_{step}$ value presented is only the initial value and is subject to change according to the attack's algorithm.}
\label{attackSetup}
\end{table}
The number of iterations given for each attack varies depending on their computational complexity and how quickly they converge. Thus, MAGE has the greatest number of iterations since the added complexity of attacking two models simultaneously requires additional attack steps to converge. MIME was given the least attack steps due to the computational cost of the attack. MIM was also given the least number of attack steps to allow for a direct comparison against MIME.

For each attack we first chose a class-wise balanced set of 1000 clean images from the testing set, of each respective data set. From this subset, we generated 1000 adversarial examples for each attack. We used 800 of these samples to create the payoff matrix $R$, then evaluated the ensemble using the remaining 200, class-wise balanced samples from each attack.

\subsection{Study of Computational Cost}
\label{sec:compS}
As the value of $n$ increases in a $\text{GaME}_n$ defense, the number of possible choices for the defender grows in accordance to the binomial coefficient, ${|D| \choose n}$, where $D$ is the set of all single model classifiers. Due to this, we provide a brief study on the effect of $n$ on the computational complexity of creating a $\text{GaME}_n$ defense.

\begin{figure}
\centering
\subfloat[Time cost as a function of n]{%
  \includegraphics[clip,width=.9\columnwidth]{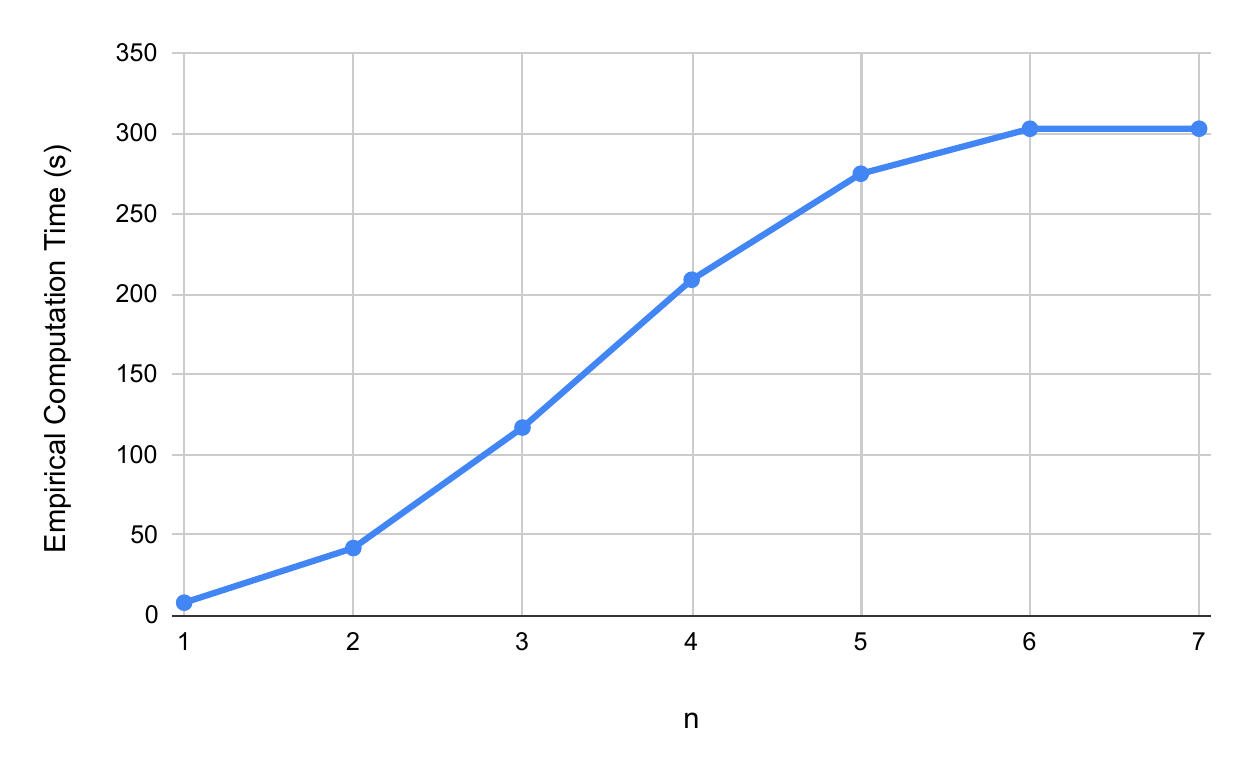}%
}

\subfloat[Time cost per model]{%
  \includegraphics[clip,width=0.9\columnwidth]{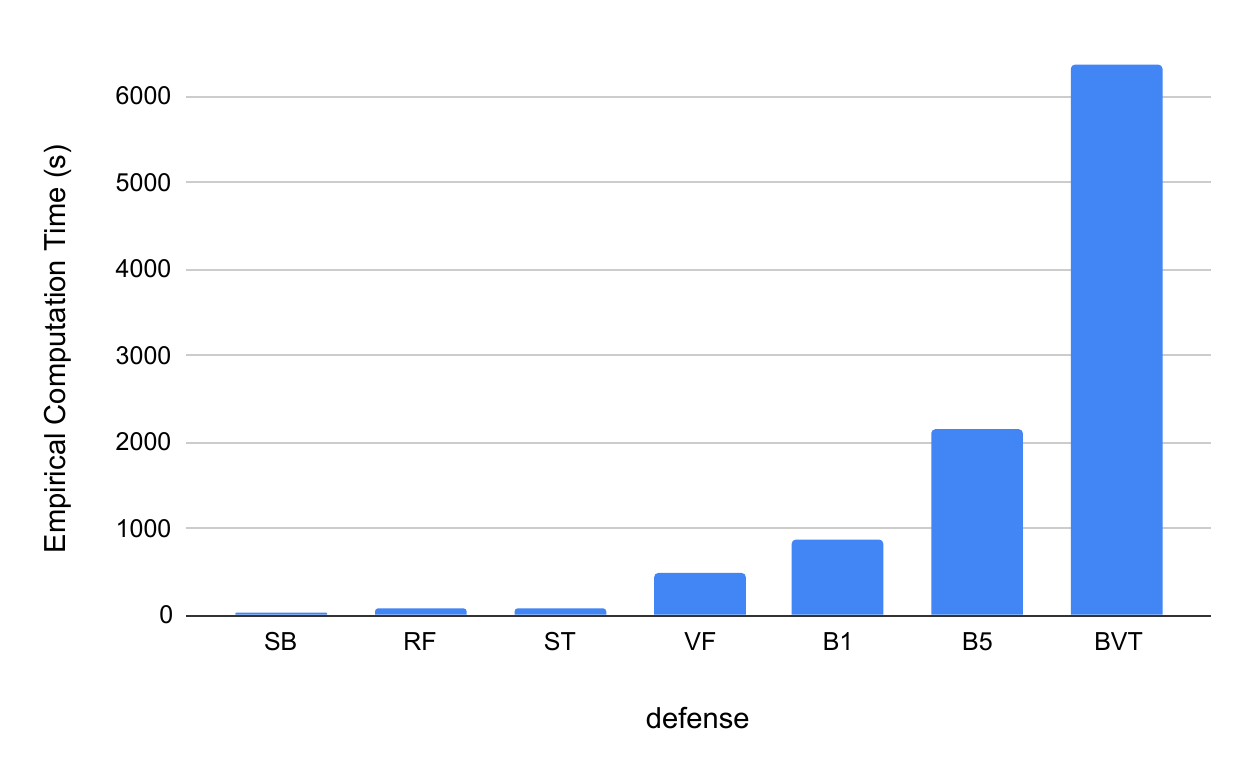}%
}
\caption{Study of the computational cost of the GaME framework (CIFAR-10). These experiments were run on a computer with the following specifications: Intel Core i9-10900K CPU @ 3.70GHz, Nvidia RTX3080 12Gb, and 64Gb RAM.}
\label{timebar}
\end{figure}

The computation time for forming the game-matrix largely depends on the time needed to compute the predictions of each of the defenses for each of the attacks, as seen in Figure~\ref{timebar}.b. Thus, rather than recomputing defense predictions when evaluating each ensemble, we can instead run each set of samples, $s_i$, through each defense $d \in D$ once, receiving output $y_{i,d}$ for each sample, defense pair. To get the robust accuracy of $U\subset D$ when evaluating samples $s_i$ we can substitute $y_{i,d}$ for $d(s_i)$ in the computation of $F_h(s_i,U)$ or $F_s(s_i,U)$ Equation~\ref{eq:fh} Equation~\ref{eq:fs}. This means that we do not need to perform a number of model evaluations that scales with $n$.

In Figure~\ref{timebar}.a we show the computational cost of creating the game matrix and solving the associated linear program for a $\text{GaME}_n$ ensemble as a function of $n$. Model prediction time is not considered in the calculation as it does not depend on $n$, as explained previously. In Figure~\ref{timebar}.b we show the computational cost of evaluating every single-model defense in the study on all 22,400 adversarial samples generated for creating the game matrix of a $\text{GaME}_n$ ensemble.The total computation time amounts to 2.9 hours of which 3\% of the time, 5 minutes, is spent evaluating each defense subset, as described above, and solving the game matrix for $\text{GaME}_7$. Both experiments were done on CIFAR-10. These experiments were run on a computer with the following specifications: Intel Core i9-10900K CPU @ 3.70GHz, Nvidia RTX3080 12Gb, and 64Gb RAM.

The computational cost for the random transform defenses were significantly greater than defenses that make a direct prediction due to the added complexity of transforming the input images. For the BaRT defenses this was mitigated by running the CPU based transformations in parallel on a per-sample basis. The computational cost for evaluating the BiT-ViT Trash is Treasure defense is the highest since it requires running a 13 step PGD attack against a large BiT model for each sample before it is given to the main ViT classifier. This transformation is also run in parallel on the GPU.

\subsection{Study on the Effects of Sample Number for GaME}
\label{sec:sampleS}
\begin{figure}
    \centering
    \includegraphics[width = \linewidth]{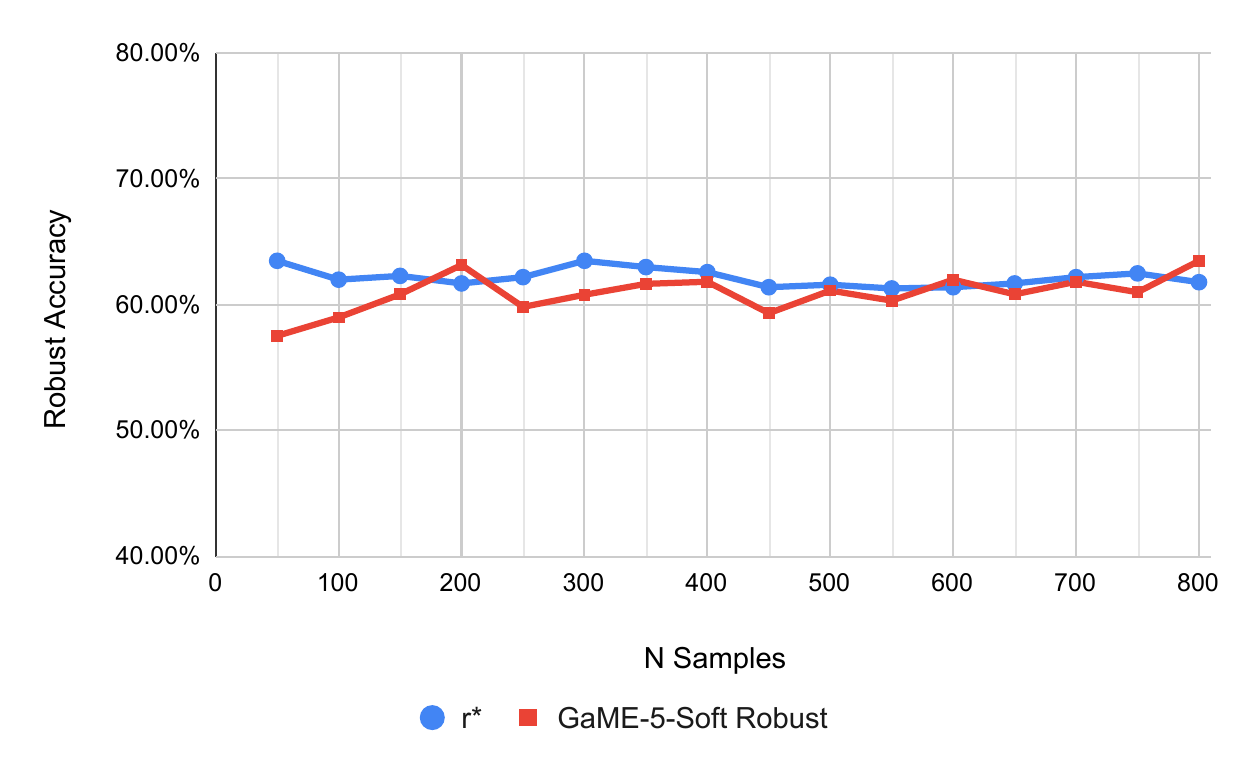}
    \caption{(CIFAR-10) Study on the effect of $N$ on the expected robust accuracy, $r^*$, in blue, and the empirical robust accuracy, in red. Results are averaged over 3 trials and measured with respect to 200 samples from each attack in the study, as described in Section~\ref{sec:setup}.}
    \label{fig:samples}
\end{figure}

In Figure~\ref{fig:samples} we provide a brief study on the effects of $N$, the number of adversarial examples generated for each attack, on the effectiveness of a GaME defense. We choose $\text{GaME}_5$ with voting function $F_s$ since it had the highest robust accuracy in our experiments. For most values of $N$ we analyzed the difference between $r^*$ and the empirical robustness was very small, in fact the average difference over all $N$ was only 1.73\%. The results show that one can achieve reasonable results with $N$ as low as 150 on CIFAR-10. This implies that one can create a GaME ensemble defense without the large computational cost of running dozens of attacks against the defenses. However, whether this trend holds for more complicated data sets with a larger number of classes such as ImageNet or even CIFAR-100 is still an open question.

\subsection{Comparison between MAGE and MAGE-D}
\label{sec:MAGE}
\begin{table}[h]
\resizebox{\linewidth}{!}{
\begin{tabular}{|c|cc|ccc|}
\hline
Dataset  & Model1     & Model2     & Clean   & MAGE    & MAGE-D  \\ \hline
CIFAR-10 & ResNet-164 & BiT-M-50x1 & 97.00\% & 60.40\% & 10.40\% \\
Tiny ImageNet    & ViT-L-16   & BiT-M-50x1 & 69.70\% & 85.30\% & 0.00\%  \\ \hline
\end{tabular}}
\caption{Comparison between MAGE and MAGE-D when attacking detector models utilizing vanilla voting models. The MAGE and MAGE-D columns represent the robust accuracy of the detector defense when evaluating samples generated by the MAGE and MAGE-D defenses respectively.}
\label{mmd}
\end{table}
MAGE-D makes significant improvements over MAGE when attacking defenses utilizing detectors as shown in Table~\ref{mmd}. For instance, when attacking the Tiny ImageNet detector, 83\% of the samples generated by MAGE are either correctly classified or labeled as adversarial by the detector, but with MAGE-D the detector's robustness drops to 0\%. This shows that MAGE-D is able to effectively break detector defense architectures by running two additional targeted attacks on image. 

\subsection{MAGE Against 3 or More Defenses}
\label{sec:threeS}
For our implementation of $\text{GaME}_n$ we performed MAGE attacks against 2 model ensembles. This was for two reasons: MAGE can have a high computational cost such that attacking size 3 ensembles increases the total number of experiments needed exponentially, and second, MAGE does not scale well to more than 2 defenses. We provide an empirical evidence of this below in Table~\ref{tbl:three} where we attack a vanilla ResNet-164 (R), vanilla ViT-L-16 (V), and a vanilla SNN model trained using back propagation.

\begin{table}[h]
\centering
\resizebox{.45\textwidth}{!}{
\begin{tabular}{|c|cccc|}
\hline
\multicolumn{1}{|c|}{Attack} & \multicolumn{1}{c}{R} & \multicolumn{1}{c}{V} & \multicolumn{1}{c}{SB} & \multicolumn{1}{c|}{R+V+SB} \\ \hline
M(R,V,SB)                 & 63.3\%                   & 31.6\%                    & 45.5\%                   & 54.6\%                           \\ \hline
\end{tabular}}
\caption{Experiment testing the effectiveness of MAGE against a defense with 3 vanilla classifiers. Each column represents the robust accuracy of the respective model or defense when evaluating the samples generated by MAGE when attacking all three models. The last column represents the robust accuracy of an ensemble defense utilizing all three models and the $f^s$ voting scheme.}
\label{tbl:three}
\end{table}

\subsection{Linear Programs for Solving GaME}
\label{sec:lpS}
We present the explicit linear program for solving GaME as the attacker. Let $O_A = (\lambda^A_{a_1}, \cdots \lambda^A_{a_{|A|}}, r^*)$ be the row vector containing the elements of $\lambda^A$ and $r^*$. Additionally let $\hat{0}$ denote the zero vector. The attacker must then solve the following linear program:
\begin{equation}
    {\small
    \begin{array}{c}
        \max \; \left( \begin{array}{c c c c}  
                0 & \cdots & 0 & 1
            \end{array} \right) O_A^T \\
            
        \text{Subject to: } \; \left( \begin{array}{c c c c}  
                -r_{d_1, a_1} & -r_{d_1, a_2} & \cdots & 1\\
                -r_{d_2, a_1} & -r_{d_2, a_2} & \cdots & 1\\
                \vdots & \vdots & \ddots & \vdots \\
                1 & 1 & \cdots & 0
            \end{array} \right) O_A^T \leq\left( \begin{array}{c}  
                0 \\
                0 \\
                \vdots \\
                1
            \end{array} \right) \\
            \text{and: }O_A \geq \hat{0}
    \end{array}}
\end{equation}
Next, we show the explicit linear program for solving GaME as the defender. For convenience let $O_D = (\lambda^D_{d_1}, \cdots \lambda^D_{d_{|D|}}, r^*)$ be the row vector containing the elements of $\lambda^D$ and $r^*$. The defender must solve the following linear program:
\begin{equation}
    {\small
    \begin{array}{c}
        \max \; \left( \begin{array}{c c c c}  
                0 & \cdots & 0 & 1
            \end{array} \right) O_D^T \\
            
        \text{Subject to: } \; \left( \begin{array}{c c c c}  
                -r_{d_1, a_1} & -r_{d_2, a_1} & \cdots & 1\\
                -r_{d_1, a_2} & -r_{d_2, a_2} & \cdots & 1\\
                \vdots & \vdots & \ddots & \vdots \\
                1 & 1 & \cdots & 0
            \end{array} \right) O_D^T \leq\left( \begin{array}{c}  
                0 \\
                0 \\
                \vdots \\
                1
            \end{array} \right) \\
            \text{and: }O_D \geq \hat{0}
    \end{array}}
\end{equation}

Due to the nature of the dual problem in linear programming, solving this problem will result in the same value for $r^*$ as was found in the primal problem presented in the main body of the paper.

\subsection{Approximating Momentum Iterative Method Over Expectation Attack}
\label{sec:approxS}
In the main body of the paper we represent the update term for the MIME attack as follows:
\begin{equation}
    g^{(i)} := \gamma g^{(i-1)} + \mathbb{E}_{t \sim T}[\frac{\partial L}{\partial t(x^{(i)}_{adv})}]
\end{equation}
In practice $g^{(i)}$ is approximated using $N$ Monte Carlo samples per input $x$:
\begin{equation}
    g^{(i)} \approx \gamma g^{(i-1)} + (\frac{1}{N} \displaystyle\sum_{j=0}^N \frac{\partial L}{\partial t_j(x^{(i)}_{adv})})
\end{equation}

\subsection{Transferability Experiments}
\label{sec:transS}
\begin{table*}[!h]
\centering
\resizebox{.95\textwidth}{!}{
\begin{tabular}{|cccccccccc|}
\hline
\multicolumn{10}{|c|}{Transferability Between Defenses (CIFAR-10)}                                                       \\ \hline
\multicolumn{1}{|c|}{Attacked} & B1      & B5      & B10     & RF      & VF      & ST      & SB      & BVT     & VRT     \\ \hline
\multicolumn{1}{|c|}{B1}       & -       & 45.80\% & 67.20\% & 99.60\% & 99.20\% & 84.10\% & 93.10\% & 95.90\% & 89.90\% \\
\multicolumn{1}{|c|}{B5}       & 4.30\%  & -       & 44.40\% & 98.70\% & 97.90\% & 71.30\% & 90.50\% & 89.70\% & 89.00\% \\
\multicolumn{1}{|c|}{B10}      & 18.60\% & 40.00\% & -       & 83.20\% & 90.80\% & 66.50\% & 77.50\% & 82.50\% & 70.20\% \\
\multicolumn{1}{|c|}{RF}       & 91.80\% & 88.40\% & 82.20\% & -       & 87.90\% & 68.60\% & 69.80\% & 80.20\% & 66.60\% \\
\multicolumn{1}{|c|}{VF}       & 52.40\% & 55.70\% & 55.70\% & 86.00\% & -       & 63.00\% & 62.90\% & 21.70\% & 74.00\% \\
\multicolumn{1}{|c|}{ST}       & 91.50\% & 90.50\% & 89.40\% & 98.90\% & 98.90\% & -       & 91.80\% & 91.80\% & 91.30\% \\
\multicolumn{1}{|c|}{SB}       & 92.20\% & 89.10\% & 86.30\% & 93.90\% & 95.20\% & 64.90\% & -       & 82.90\% & 84.70\% \\
\multicolumn{1}{|c|}{BVT}      & 60.40\% & 69.90\% & 75.10\% & 98.20\% & 95.60\% & 78.10\% & 90.00\% & -       & 90.80\% \\
\multicolumn{1}{|c|}{VRT}      & 86.80\% & 83.40\% & 89.80\% & 82.30\% & 87.20\% & 76.00\% & 71.00\% & 78.80\% & -       \\ \hline
\end{tabular}}
\caption{Full numerical results for the transfer study shown pictorially in the main body of the paper. Here each value represents the robust accuracy of the column defense when evaluating samples generated by attacking the row defense. Thus, smaller values represent a higher level of transferability since more samples are incorrectly classified by the evaluating, column defense.}
\label{tbl:transferabilit2}
\end{table*}

Previously, we gave a graphical representation of the transferability results for CIFAR-10. In Table~\ref{tbl:transferabilit2}, we include the full results from our transferability study. Each of the defense names have been abbreviated in the interest of space: Bn is BaRT-n, RF is the FAT trained ResNet-164, VF is the FAT trained ViT-L, ST is the transfer SNN, SB is the Back-Prop SNN, VRT is the Vgg + ResNet TiT defense, and BVT is the ViT+BiT TiT defense.

We first chose 1000 class-wise balanced, clean images from the testing set of CIFAR-10. We additionally constrained these 1000 samples to be those which are correctly identified by every model in the table. For random transform defenses, each image is classified correctly with a probability of at least 98\%. We then attacked each defense with APGD, every randomized defense with MIME, and every non-randomized defense with MIM. For APGD, MIM, and MIME the parameters can be seen in Subsection~\ref{sec:setup}. From each defense we chose the adversarial samples generated by the attack with the highest attack success rate to be used in the transfer study.

\textbf{Analysis of Results}
From the transferability table it becomes clear that there is generally very low transferability between each pair of defenses which use different classifier architectures. For instance, attacks generated on the BiT based BaRT models have a very low level of transferability with the ViT based FAT-ViT and TiT defense using BiT+ViT, with the maximum transferability being 82.5\%. In contrast to this, defenses which share architectures, unsurprisingly, have relatively high levels of transferability. One notable example of this is the attacks generated on the FAT-ViT models when transfered to the TiT defense using BiT-ViT, acheiving a high transferability of 21.7\%. There are some exceptions to this idea however, as the BaRT models seem to be highly susceptible to attacks generated on the ViT based defenses. For instance, samples generated on the FAT trained ViT-L model have a relatively high level of transferability to the BaRT models, acheiving 52.4\%, 55.7\%, and 55.7\% for BaRT-1,5 and 10 respectively.

\subsection{Full Experimental Results for GaME}
\label{sec:resultsS}

\begin{table*}
	\begin{minipage}{0.5\linewidth}
		\centering
		\includegraphics[width=\linewidth]{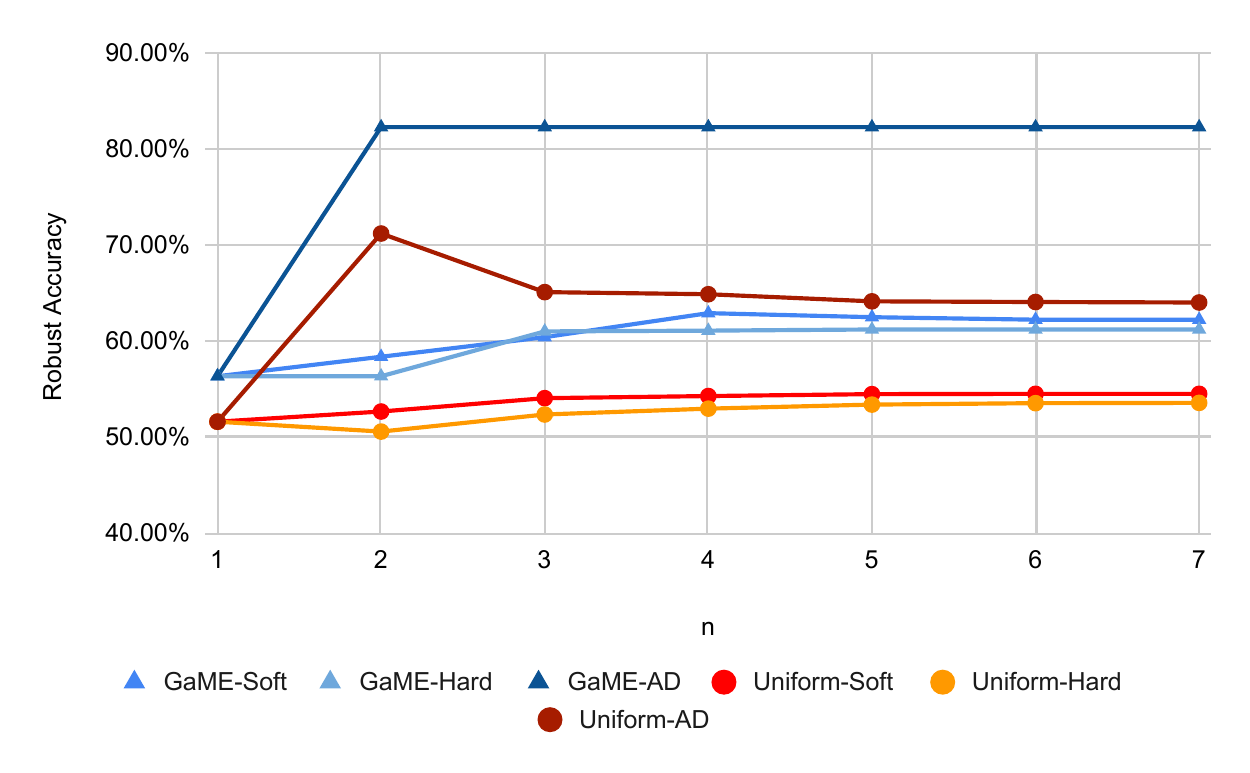}
	\end{minipage}\hfill
	\begin{minipage}{0.45\linewidth}
		\centering
		\includegraphics[width=\linewidth]{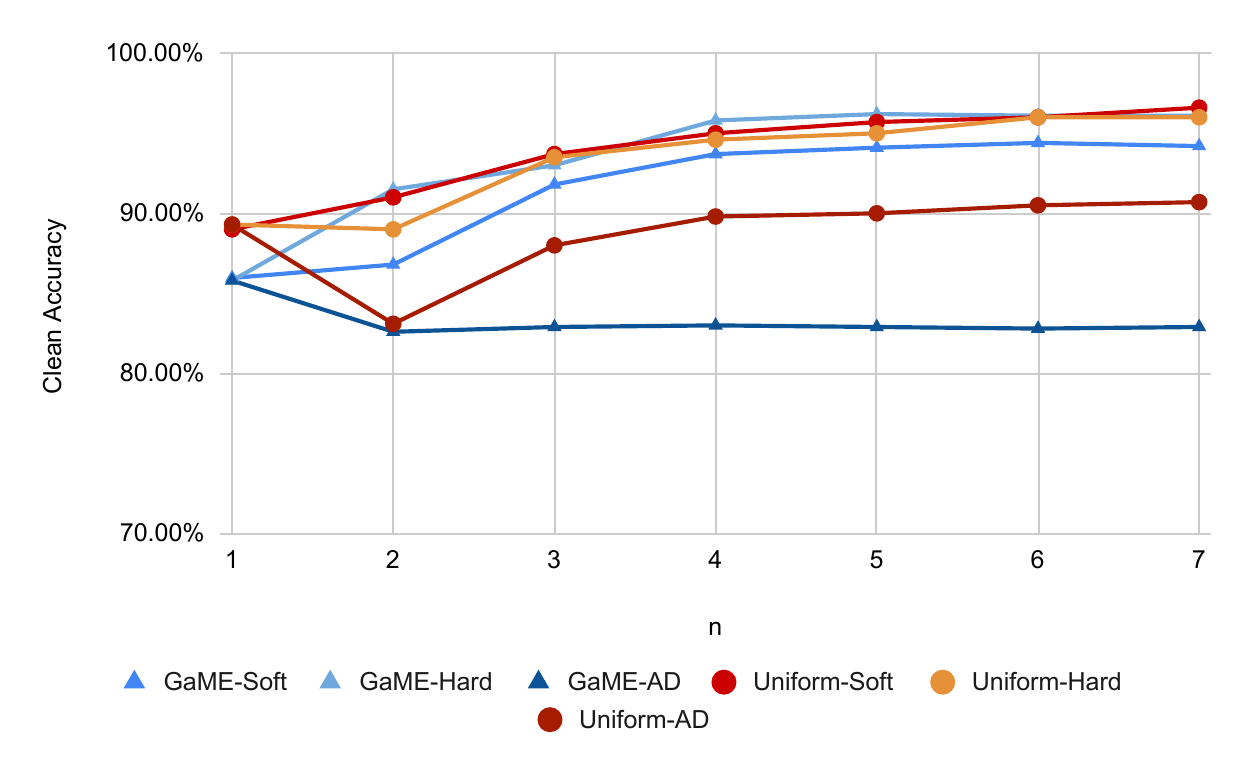}
	\end{minipage}
    \captionsetup{type=figure}
    \caption{(CIFAR-10) Comparison between GaME generated ensemble defenses, and ensemble defenses with uniform probability distributions over all strategies. Tests are done for all possible values of n, the largest ensemble size.}
    \label{fig:robcifar}
\end{table*}
\begin{table*}
	\begin{minipage}{0.5\linewidth}
		\centering
		\includegraphics[width=\linewidth]{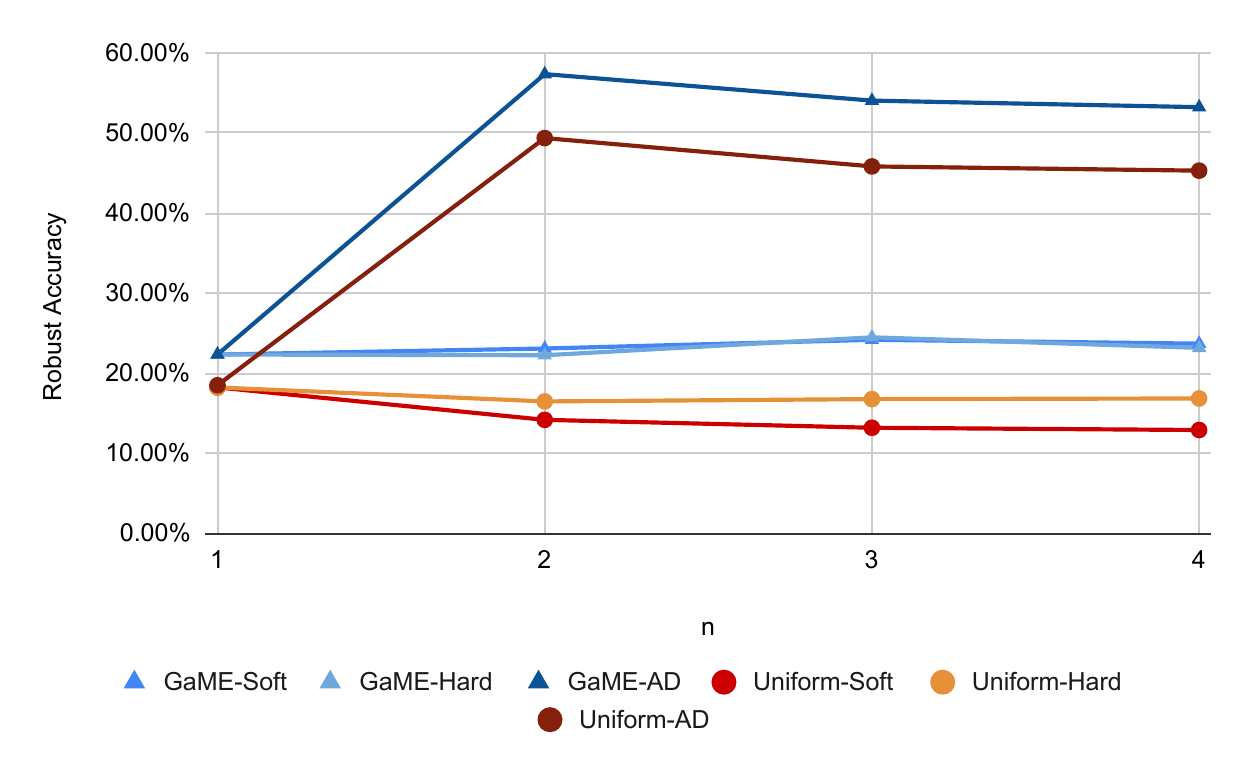}
	\end{minipage}\hfill
	\begin{minipage}{0.45\linewidth}
		\centering
		\includegraphics[width=\linewidth]{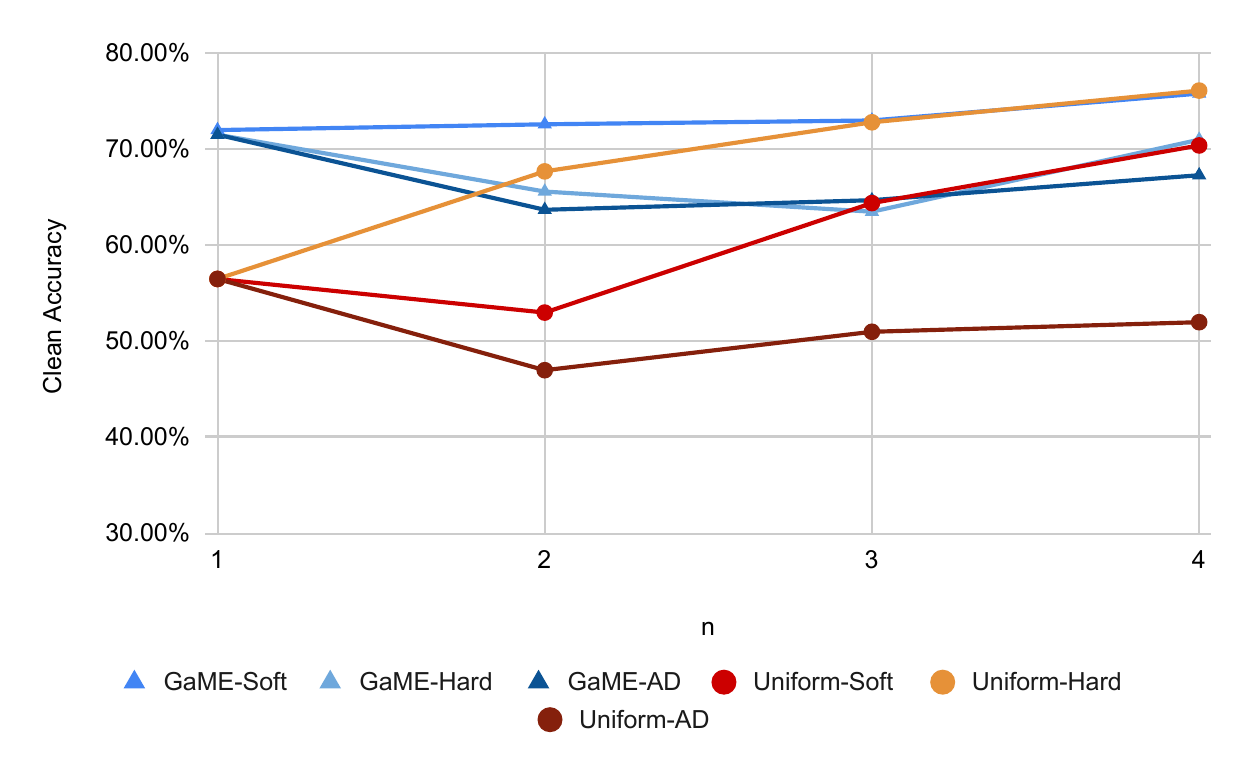}
	\end{minipage}
    \captionsetup{type=figure}
    \caption{(Tiny ImageNet) Comparison between GaME generated ensemble defenses, and ensemble defenses with uniform probability distributions over all strategies. Tests are done for all possible values of n, the largest ensemble size.}
    \label{fig:robtiny}
\end{table*}

In Figure~\ref{fig:robcifar} and Figure~\ref{fig:robtiny} we compare the robust and clean accuracy for GaME generated ensemble defenses and those with uniform probability distributions over all defender strategies. We study all values of $1\leq n \leq |D|$ for each dataset. The results show that the GaME framework clearly improves upon the robustness of the uniform probability distribution ensemble while maintaining a high level of clean accuracy. The results additionally show that the clean and robust accuracy both generally trend in a positive direction as one increases $n$. The full numerical results for these figures can be found in Table~\ref{tbl:c10full} and Table~\ref{tbl:tinyfull}. The notation for these tables and graphs are as follows: column titles starting with "GaMe" use mixed strategies generated by the GaME framework, while those starting with "Uniform" use probability distributions that are uniform over all strategies; those ending with "Hard", "Soft", and "AD" use the $f^h$, $f^s$, and $f^d$ voting schemes respectively.

In Table~\ref{tbl:GaME1} we show the mixed Nash strategies for the defender in $\text{GaME}_1$ on CIFAR-10 and Tiny ImageNet. We do the same in Table~\ref{tblA:GaME1} for the mixed Nash strategy of the attacker. In Table~\ref{tbl:c10game} and Table~\ref{tbl:tinygame} we give the full utility matrices for the defender in our instances of $\text{GaME}_1$ on CIFAR-10 and Tiny ImageNet respectively. Here we use M(A,B) and MD(A,B) to denote the MAGE and MAGE-D attacks run against defenses A and B.

The results in Table~\ref{tblA:GaME1} show that the best strategies for the attacker tend to implement the multi-model MAGE and MAGE-D attacks. It is worth noting that in each of these cases the two targeted defenses utilize different architectures. This result aligns well with the results presented in subsection~\ref{sec:transS}, as defenses with a shared architecture tend to have a high level of transferability. Thus, by targeting two different base architectures, the attacker can effectively transfer the attack to many more defenses in the ensemble. This is in contrast to attacking two models with a shared architecture which will likely result in adversarial examples with a low level of transferability to different architectures.
\begin{table*}
\centering
\resizebox{\linewidth}{!}{
\begin{tabular}{|ccccccc|cccccc|}
\hline
\multicolumn{7}{|c|}{Robust}                                                                         & \multicolumn{6}{c|}{Clean}                                                 \\ \hline
\multicolumn{1}{|c|}{n} & GaME-Soft & GaME-Hard & GaME-AD  & Uniform-Soft & Uniform-Hard & Uniform-AD & GaME-Soft & GaME-Hard & GaME-AD & Uniform-Soft & Uniform-Hard & Uniform-AD \\ \hline
\multicolumn{1}{|c|}{1} & 56.37\%   & 56.37\%   & 56.37\% & 51.64\%      & 51.64\%      & 51.64\%    & 85.96\%   & 85.80\%   & 85.80\% & 89.00\%      & 89.30\%      & 89.30\%    \\
\multicolumn{1}{|c|}{2} & 58.40\%   & 56.37\%   & 82.31\% & 52.70\%      & 50.61\%      & 71.23\%    & 86.80\%   & 91.50\%   & 82.60\% & 91.00\%      & 89.00\%      & 83.10\%    \\
\multicolumn{1}{|c|}{3} & 60.44\%   & 61.04\%   & 82.31\% & 54.09\%      & 52.39\%      & 65.13\%    & 91.80\%   & 93.00\%   & 82.90\% & 93.70\%      & 93.50\%      & 88.00\%    \\
\multicolumn{1}{|c|}{4} & 62.94\%   & 61.12\%   & 82.31\% & 54.31\%      & 53.00\%      & 64.91\%    & 93.70\%   & 95.80\%   & 83.00\% & 95.00\%      & 94.60\%      & 89.80\%    \\
\multicolumn{1}{|c|}{5} & 62.52\%   & 61.24\%   & 82.31\% & 54.51\%      & 53.42\%      & 64.17\%    & 94.10\%   & 96.20\%   & 82.90\% & 95.70\%      & 95.00\%      & 90.00\%    \\
\multicolumn{1}{|c|}{6} & 62.26\%   & 61.24\%   & 82.31\% & 54.54\%      & 53.57\%      & 64.10\%    & 94.40\%   & 96.10\%   & 82.80\% & 96.00\%      & 96.00\%      & 90.50\%    \\
\multicolumn{1}{|c|}{7} & 62.26\%   & 61.24\%   & 82.31\% & 54.54\%      & 53.60\%      & 64.05\%    & 94.20\%   & 96.10\%   & 82.90\% & 96.60\%      & 96.00\%      & 90.70\%    \\ \hline
\end{tabular}}
\caption{(CIFAR-10) Full numerical results for \ref{fig:robcifar}. Here we can see that the GaME generated ensembles are able to out perform the uniform distribution ensembles while maintaining improved or similar levels of clean accuracy. For instance, the GaME-AD defense improves upon the robustness of the Uniform-AD (n=2) defense by 11.08\%, while experiencing a drop in clean accuracy of only .05\%.}
\label{tbl:c10full}
\end{table*}
\begin{table*}
\centering
\resizebox{\linewidth}{!}{
\begin{tabular}{|ccccccc|cccccc|}
\hline
\multicolumn{7}{|c|}{Robust}                                                                        & \multicolumn{6}{c|}{Clean}                                                 \\ \hline
\multicolumn{1}{|c|}{n} & GaME-Soft & GaME-Hard & GaME-AD & Uniform-Soft & Uniform-Hard & Uniform-AD & GaME-Soft & GaME-Hard & GaME-AD & Uniform-Soft & Uniform-Hard & Uniform-AD \\ \hline
\multicolumn{1}{|c|}{1} & 22.36\%    & 22.36\%   & 22.36\% & 18.25\%      & 18.25\%       & 18.52\%   & 72.00\%   & 71.52\%   & 71.52\% & 56.50\%      & 56.50\%      & 56.50\%    \\
\multicolumn{1}{|c|}{2} & 23.11\%    & 22.27\%   & 57.39\% & 14.2\%      & 16.5\%       & 49.38\%    & 72.60\%   & 65.60\%   & 63.70\% & 53.00\%      & 67.70\%      & 47.00\%    \\
\multicolumn{1}{|c|}{3} & 24.2\%    & 24.51\%   & 54.08\% & 13.21\%     & 16.79\%     & 45.86\%    & 73.00\%   & 63.50\%   & 64.70\% & 64.40\%      & 72.81\%      & 51.00\%    \\
\multicolumn{1}{|c|}{4} & 23.72\%   & 23.19\%   & 53.26\% & 12.93\%      & 16.87\%     & 45.33\%    & 75.80\%   & 71.00\%   & 67.30\% & 70.40\%      & 76.11\%      & 52.00\%    \\ \hline
\end{tabular}}
\caption{(Tiny ImageNet) Full numerical results for \ref{fig:robtiny}. Here we can again see that the GaME generated ensembles are able to out perform the uniform distribution ensembles while maintaining improved or similar levels of clean accuracy. Similarly to CIFAR-10, the GaME-AD defense improves upon the robustness of the Uniform-AD (n=2) defense by 8.01\%, but also improves upon the clean accuracy by 16.7\%.}
\label{tbl:tinyfull}
\end{table*}\newpage
\begin{table*}
      \centering
\begin{tabular}{|c|ccccccc|}
\hline \multicolumn{8}{|c|}{Defender Mixed Nash Strategy (CIFAR-10, $\text{GaME}_1$)}   \\ \hline
        Defense    & B1  & B5   & RF   & VF   & ST  & SB  & BVT \\ \hline
$\lambda^D$ & 0   & .236  & .365  & .035  & .270   & .094 & 0   \\ \hline
Min Robust  & 2.5\% & 17\% & 45\% & 28.5\% & 0\% & 1\% & 1.5\% \\ \hline\hline
\multicolumn{8}{|c|}{Defender Mixed Nash Strategy (Tiny ImageNet, $\text{GaME}_1$)} \\ \hline
$\lambda^D$ & 0.138    & 0.425  & -  & 0.304 & - & -   & 0.133   \\ \hline
Min Robust  & 2\%   & 11\% & -  & 2\% & - & -   & 1\%  \\ \hline
\end{tabular}
\caption[]{$GaME_{1}$ Defender results for CIFAR-10 and Tiny ImageNet. Here $\lambda^D$ for a particular defense represents the probability that it will be used for classification of an input image. Furthermore, min robust represents the lowest robust accuracy achieved by any attack against the given defense.}
\label{tbl:GaME1}
\end{table*}

\begin{table*}
\centering
\begin{tabular}{|cccccc|}
\hline
\multicolumn{6}{|c|}{Attacker Mixed Nash Strategy (CIFAR-10, $\text{GaME}_1$)}           \\ \hline
\multicolumn{1}{|c|}{Attack}     & M(B5,SB)  & MD(B5,RF) & MD(B5,ST) & MD(RF,VF) & MD(RF,ST)  \\ \hline
\multicolumn{1}{|c|}{$\lambda^A$}          & 0.107     & 0.212    & 0.033    & 0.624 & 0.024     \\ \hline
\multicolumn{1}{|c|}{Max Robust} & 92.5\%      & 89\%     & 93\%     & 67\%  &   92\%   \\ \hline \hline
\multicolumn{6}{|c|}{Attacker Mixed Nash Strategy (Tiny ImageNet, $\text{GaME}_1$)}      \\ \hline
\multicolumn{1}{|c|}{Attack}     & M(B1,BVT) & M(B1,VF) & MIME(B5) & M(B5,BVT) & - \\ \hline
\multicolumn{1}{|c|}{$\lambda^A$}          & 0.192     & 0.184    & 0.102    & 0.522 & -    \\ \hline
\multicolumn{1}{|c|}{Max Robust} & 36\%      & 56\%     & 62\%     & 38\%  & -    \\ \hline
\end{tabular}
 \caption[]{$GaME_{1}$ Attacker results for CIFAR-10 and Tiny ImageNet. Attacks which have a probability of 0 in the mixed strategy are not represented here. The results show that it is generally a better strategy for the attacker to implement multi-model attacks, like MAGE and MAGE-D, instead of single model attacks, like APGD, when targeting ensemble defenses. Max robust represents the highest robust accuracy achieved by any defense when evaluating samples generated by the given attack.}
\label{tblA:GaME1}
\end{table*}

\begin{table*}
\centering
\begin{tabular}{|c|cccc|}
\hline
Attack     & B1      & B5      & VF      & BVT     \\ \hline
APGD(VF)   & 53.50\% & 47.50\% & 4.50\%  & 33.00\% \\
MIME(B1)   & 6.00\%  & 42.00\% & 34.50\% & 70.50\% \\
MIME(B5)   & 27.00\% & 14.50\% & 33.50\% & 67.00\% \\
MIME(BVT)  & 58.00\% & 57.50\% & 33.50\% & 3.50\%  \\
M(B1,B5)   & 6.00\%  & 19.50\% & 32.50\% & 72.00\% \\
M(B1,BVT)  & 6.00\%  & 44.00\% & 33.50\% & 8.00\%  \\
M(B1,VF)   & 7.50\%  & 37.00\% & 9.00\%  & 59.50\% \\
M(B5,BVT)  & 42.00\% & 24.00\% & 32.50\% & 13.50\% \\
M(B5,VF)   & 47.00\% & 27.50\% & 12.00\% & 69.00\% \\
M(VF,VBT)  & 59.00\% & 51.00\% & 6.00\%  & 3.00\%  \\
MD(B1,B5)  & 4.00\%  & 16.00\% & 28.00\% & 67.00\% \\
MD(B1,BVT) & 6.00\%  & 33.00\% & 28.00\% & 6.00\%  \\
MD(B1,VF)  & 2.00\%  & 37.00\% & 8.00\%  & 56.00\% \\
MD(B5,BVT) & 37.00\% & 16.00\% & 30.00\% & 10.00\% \\
MD(B5,VF)  & 35.00\% & 18.00\% & 8.00\%  & 57.00\% \\
MD(BVT,VF) & 44.00\% & 40.00\% & 2.00\%  & 1.00\%  \\ \hline
\end{tabular}
\caption{Here we provide the full utility matrix for the defender when creating a $\text{GaME}_1$ defense on Tiny ImageNet. All values represent robust accuracy of the column defense when evaluating samples generated by the row attack.}
\label{tbl:tinygame}
\end{table*}

\begin{table*}
\centering
\begin{tabular}{|c|ccccccc|}
\hline
Attack     & B1      & B5      & RF      & VF      & ST      & SB      & BVT     \\ \hline
APGD(RF)   & 91.50\% & 87.00\% & 45.00\% & 84.00\% & 68.00\% & 59.50\% & 75.00\% \\
APGD(SB)   & 89.50\% & 86.50\% & 83.50\% & 89.50\% & 62.00\% & 1.50\%  & 74.50\% \\
APGD(ST)   & 92.00\% & 86.50\% & 86.00\% & 95.50\% & 0.00\%  & 82.00\% & 88.50\% \\
APGD(VF)   & 56.00\% & 58.00\% & 76.00\% & 28.50\% & 60.50\% & 54.50\% & 28.00\% \\
MIME(B1)   & 2.50\%  & 44.50\% & 85.50\% & 95.00\% & 81.50\% & 83.00\% & 91.00\% \\
MIME(B5)   & 6.00\%  & 17.00\% & 84.50\% & 94.00\% & 71.00\% & 78.50\% & 85.00\% \\
MIME(BVT)  & 53.50\% & 66.00\% & 86.00\% & 92.00\% & 76.00\% & 80.00\% & 12.00\% \\
M(B1,B5)   & 5.00\%  & 36.50\% & 84.50\% & 94.50\% & 80.00\% & 81.00\% & 87.00\% \\
M(B1,BVT)  & 7.00\%  & 54.50\% & 85.50\% & 94.50\% & 74.50\% & 81.50\% & 34.00\% \\
M(B1,RF)   & 11.00\% & 61.50\% & 55.00\% & 89.00\% & 74.50\% & 74.00\% & 81.00\% \\
M(B1,SB)   & 7.00\%  & 56.50\% & 81.50\% & 91.50\% & 62.50\% & 9.50\%  & 85.00\% \\
M(B1,ST)   & 16.50\% & 68.50\% & 84.50\% & 95.00\% & 0.50\%  & 82.50\% & 90.00\% \\
M(B1,VF)   & 6.00\%  & 48.50\% & 82.50\% & 73.00\% & 76.50\% & 75.50\% & 74.50\% \\
M(B5,BVT)  & 17.00\% & 33.50\% & 85.50\% & 94.50\% & 73.50\% & 81.00\% & 36.00\% \\
M(B5,RF)   & 18.50\% & 36.50\% & 55.50\% & 90.50\% & 70.50\% & 72.00\% & 83.00\% \\
M(B5,SB)   & 19.00\% & 28.50\% & 82.50\% & 92.50\% & 63.00\% & 8.50\%  & 80.50\% \\
M(B5,ST)   & 63.50\% & 60.00\% & 85.00\% & 94.50\% & 0.50\%  & 83.50\% & 88.50\% \\
M(B5,VF)   & 12.50\% & 29.50\% & 81.50\% & 71.50\% & 75.00\% & 75.50\% & 71.00\% \\
M(RF,BVT)  & 70.50\% & 77.50\% & 59.50\% & 85.00\% & 66.00\% & 71.00\% & 7.50\%  \\
M(RF,SB)   & 89.00\% & 85.50\% & 50.00\% & 84.50\% & 60.50\% & 3.00\%  & 71.00\% \\
M(RF,SBT)  & 92.50\% & 90.00\% & 73.50\% & 93.00\% & 0.00\%  & 74.00\% & 83.00\% \\
M(RF,VF)   & 66.50\% & 62.50\% & 52.00\% & 40.50\% & 62.50\% & 54.50\% & 36.00\% \\
M(SB,ST)   & 91.50\% & 90.50\% & 85.50\% & 94.50\% & 0.00\%  & 33.50\% & 84.50\% \\
M(SB,VBT)  & 82.00\% & 82.50\% & 83.50\% & 89.00\% & 62.00\% & 4.50\%  & 18.00\% \\
M(ST,BVT)  & 90.50\% & 90.00\% & 86.50\% & 95.00\% & 0.50\%  & 81.00\% & 25.50\% \\
M(VF, BVT) & 48.00\% & 60.00\% & 76.50\% & 52.00\% & 65.00\% & 68.00\% & 1.50\%  \\
M(VF,SB)   & 83.50\% & 83.00\% & 81.50\% & 79.00\% & 57.50\% & 2.00\%  & 66.50\% \\
M(VF,ST)   & 85.00\% & 81.00\% & 84.50\% & 88.50\% & 1.00\%  & 76.00\% & 78.50\% \\
MD(B1,B5)  & 8.00\%  & 37.00\% & 90.00\% & 94.00\% & 82.00\% & 87.00\% & 90.00\% \\
MD(B1,BVT) & 4.00\%  & 53.00\% & 90.00\% & 93.00\% & 72.00\% & 85.00\% & 14.00\% \\
MD(B1,RF)  & 14.00\% & 77.00\% & 67.00\% & 89.00\% & 77.00\% & 80.00\% & 88.00\% \\
MD(B1,SB)  & 10.00\% & 60.00\% & 84.00\% & 91.00\% & 60.00\% & 7.00\%  & 83.00\% \\
MD(B1,ST)  & 17.00\% & 75.00\% & 90.00\% & 93.00\% & 0.00\%  & 85.00\% & 94.00\% \\
MD(B1,VF)  & 9.00\%  & 49.00\% & 88.00\% & 70.00\% & 81.00\% & 76.00\% & 76.00\% \\
MD(B5,BVT) & 16.00\% & 29.00\% & 89.00\% & 93.00\% & 72.00\% & 87.00\% & 24.00\% \\
MD(B5,RF)  & 21.00\% & 36.00\% & 63.00\% & 89.00\% & 72.00\% & 71.00\% & 83.00\% \\
MD(B5,SB)  & 18.00\% & 37.00\% & 82.00\% & 92.00\% & 64.00\% & 9.00\%  & 84.00\% \\
MD(B5,ST)  & 63.00\% & 66.00\% & 90.00\% & 93.00\% & 0.00\%  & 83.00\% & 89.00\% \\
MD(B5,VF)  & 14.00\% & 27.00\% & 86.00\% & 77.00\% & 72.00\% & 80.00\% & 82.00\% \\
MD(RF,BVT) & 72.00\% & 77.00\% & 64.00\% & 86.00\% & 70.00\% & 74.00\% & 7.00\%  \\
MD(RF,SB)  & 87.00\% & 82.00\% & 59.00\% & 82.00\% & 56.00\% & 6.00\%  & 83.00\% \\
MD(RF,ST)  & 92.00\% & 91.00\% & 74.00\% & 90.00\% & 0.00\%  & 74.00\% & 85.00\% \\
MD(RF,VF)  & 67.00\% & 65.00\% & 58.00\% & 47.00\% & 58.00\% & 60.00\% & 33.00\% \\
MD(SB,VBT) & 83.00\% & 79.00\% & 87.00\% & 89.00\% & 61.00\% & 2.00\%  & 5.00\%  \\
MD(ST,BVT) & 83.00\% & 83.00\% & 90.00\% & 93.00\% & 0.00\%  & 86.00\% & 15.00\% \\
MD(ST,SB)  & 86.00\% & 89.00\% & 87.00\% & 92.00\% & 0.00\%  & 23.00\% & 87.00\% \\
MD(VF,BVT) & 53.00\% & 61.00\% & 88.00\% & 51.00\% & 64.00\% & 76.00\% & 2.00\%  \\
MD(VF,SB)  & 71.00\% & 73.00\% & 80.00\% & 60.00\% & 55.00\% & 1.00\%  & 57.00\% \\
MD(VF,ST)  & 79.00\% & 80.00\% & 88.00\% & 84.00\% & 2.00\%  & 80.00\% & 77.00\% \\ \hline
\end{tabular}
\caption{Here we present the full results from $\text{GaME}_1$ performed on CIFAR-10. This extensive study shows us that, for any attack, there is going to be a defense that can counter it, and vice versa. For instance, MIME(B5) is extremely effective against BaRT-1 and BaRT-5, yet it fails to fool the ViT-FAT model 94\% of the time.  All values represent robust accuracy of the column defense when evaluating samples generated by the row attack.}
\label{tbl:c10game}
\end{table*}


%% file: egpaper_final.bbl
\begin{thebibliography}{10}\itemsep=-1pt

\bibitem{abnar2020quantifying}
Samira Abnar and Willem Zuidema.
\newblock Quantifying attention flow in transformers.
\newblock In {\em Proceedings of the 58th Annual Meeting of the Association for
  Computational Linguistics}, pages 4190--4197, 2020.

\bibitem{araujo2020advocating}
Alexandre Araujo, Laurent Meunier, Rafael Pinot, and Benjamin Negrevergne.
\newblock Advocating for multiple defense strategies against adversarial
  examples.
\newblock In {\em Joint European Conference on Machine Learning and Knowledge
  Discovery in Databases}, pages 165--177. Springer, 2020.

\bibitem{EOT}
Anish Athalye, Logan Engstrom, Andrew Ilyas, and Kevin Kwok.
\newblock Synthesizing robust adversarial examples.
\newblock In {\em International conference on machine learning}, pages
  284--293. PMLR, 2018.

\bibitem{balcan2022nash}
Maria-Florina Balcan, Rattana Pukdee, Pradeep Ravikumar, and Hongyang Zhang.
\newblock Nash equilibria and pitfalls of adversarial training in adversarial
  robustness games.
\newblock {\em arXiv preprint arXiv:2210.12606}, 2022.

\bibitem{bypassing10detectors}
Nicholas Carlini and David Wagner.
\newblock Adversarial examples are not easily detected: Bypassing ten detection
  methods.
\newblock In {\em Proceedings of the 10th ACM workshop on artificial
  intelligence and security}, pages 3--14, 2017.

\bibitem{APGD}
Francesco Croce and Matthias Hein.
\newblock Reliable evaluation of adversarial robustness with an ensemble of
  diverse parameter-free attacks.
\newblock In {\em International conference on machine learning}, pages
  2206--2216. PMLR, 2020.

\bibitem{MIM}
Yinpeng Dong, Fangzhou Liao, Tianyu Pang, Hang Su, Jun Zhu, Xiaolin Hu, and
  Jianguo Li.
\newblock Boosting adversarial attacks with momentum.
\newblock In {\em Proceedings of the IEEE Conference on Computer Vision and
  Pattern Recognition (CVPR)}, June 2018.

\bibitem{VIT}
Alexey Dosovitskiy, Lucas Beyer, Alexander Kolesnikov, Dirk Weissenborn,
  Xiaohua Zhai, Thomas Unterthiner, Mostafa Dehghani, Matthias Minderer, Georg
  Heigold, Sylvain Gelly, et~al.
\newblock An image is worth 16x16 words: Transformers for image recognition at
  scale.
\newblock In {\em International Conference on Learning Representations}, 2020.

\bibitem{SEW}
Wei Fang, Zhaofei Yu, Yanqi Chen, Tiejun Huang, Timoth{\'e}e Masquelier, and
  Yonghong Tian.
\newblock Deep residual learning in spiking neural networks.
\newblock {\em Advances in Neural Information Processing Systems},
  34:21056--21069, 2021.

\bibitem{medicineAdvML}
Samuel~G Finlayson, John~D Bowers, Joichi Ito, Jonathan~L Zittrain, Andrew~L
  Beam, and Isaac~S Kohane.
\newblock Adversarial attacks on medical machine learning.
\newblock {\em Science}, 363(6433):1287--1289, 2019.

\bibitem{goodfellow2014explaining}
Ian~J Goodfellow, Jonathon Shlens, and Christian Szegedy.
\newblock Explaining and harnessing adversarial examples.
\newblock {\em arXiv preprint arXiv:1412.6572}, 2014.

\bibitem{BIT}
Alexander Kolesnikov, Lucas Beyer, Xiaohua Zhai, Joan Puigcerver, Jessica Yung,
  Sylvain Gelly, and Neil Houlsby.
\newblock Big transfer (bit): General visual representation learning.
\newblock In {\em European conference on computer vision}, pages 491--507.
  Springer, 2020.

\bibitem{c10}
Alex Krizhevsky, Vinod Nair, and Geoffrey Hinton.
\newblock Cifar-10 (canadian institute for advanced research).

\bibitem{pmlr-v151-le22c}
Trung Le, Anh Tuan~Bui, Le Minh Tri~Tue, He Zhao, Paul Montague, Quan Tran, and
  Dinh Phung.
\newblock On global-view based defense via adversarial attack and defense risk
  guaranteed bounds.
\newblock In Gustau Camps-Valls, Francisco J.~R. Ruiz, and Isabel Valera,
  editors, {\em Proceedings of The 25th International Conference on Artificial
  Intelligence and Statistics}, volume 151 of {\em Proceedings of Machine
  Learning Research}, pages 11438--11460. PMLR, 28--30 Mar 2022.

\bibitem{le2015tiny}
Ya Le and Xuan Yang.
\newblock Tiny imagenet visual recognition challenge.
\newblock {\em CS 231N}, 7(7):3, 2015.

\bibitem{Delving}
Yanpei Liu, Xinyun Chen, Chang Liu, and Dawn Song.
\newblock Delving into transferable adversarial examples and black-box attacks.
\newblock {\em arXiv preprint arXiv:1611.02770}, 2016.

\bibitem{madry2018towards}
Aleksander Madry, Aleksandar Makelov, Ludwig Schmidt, Dimitris Tsipras, and
  Adrian Vladu.
\newblock Towards deep learning models resistant to adversarial attacks.
\newblock In {\em International Conference on Learning Representations}, 2018.

\bibitem{mahmood2021beware}
Kaleel Mahmood, Deniz Gurevin, Marten van Dijk, and Phuoung~Ha Nguyen.
\newblock Beware the black-box: On the robustness of recent defenses to
  adversarial examples.
\newblock {\em Entropy}, 23(10):1359, 2021.

\bibitem{mahmood2021robustness}
Kaleel Mahmood, Rigel Mahmood, and Marten Van~Dijk.
\newblock On the robustness of vision transformers to adversarial examples.
\newblock In {\em Proceedings of the IEEE/CVF International Conference on
  Computer Vision}, pages 7838--7847, 2021.

\bibitem{BARZ}
Kaleel Mahmood, Phuong~Ha Nguyen, Lam~M. Nguyen, Thanh Nguyen, and Marten
  Van~Dijk.
\newblock Besting the black-box: Barrier zones for adversarial example defense.
\newblock {\em IEEE Access}, 10:1451--1474, 2022.

\bibitem{maini2020adversarial}
Pratyush Maini, Eric Wong, and Zico Kolter.
\newblock Adversarial robustness against the union of multiple perturbation
  models.
\newblock In {\em International Conference on Machine Learning}, pages
  6640--6650. PMLR, 2020.

\bibitem{meunier2021mixed}
Laurent Meunier, Meyer Scetbon, Rafael~B Pinot, Jamal Atif, and Yann
  Chevaleyre.
\newblock Mixed nash equilibria in the adversarial examples game.
\newblock In {\em International Conference on Machine Learning}, pages
  7677--7687. PMLR, 2021.

\bibitem{Nash}
John Nash.
\newblock Non-cooperative games.
\newblock {\em Annals of Mathematics}, 54(2):286--295, 1951.

\bibitem{pal2020game}
Ambar Pal and Ren{\'e} Vidal.
\newblock A game theoretic analysis of additive adversarial attacks and
  defenses.
\newblock {\em Advances in Neural Information Processing Systems},
  33:1345--1355, 2020.

\bibitem{pang2019improving}
Tianyu Pang, Kun Xu, Chao Du, Ning Chen, and Jun Zhu.
\newblock Improving adversarial robustness via promoting ensemble diversity.
\newblock In {\em International Conference on Machine Learning}, pages
  4970--4979. PMLR, 2019.

\bibitem{papernot2016transferability}
Nicolas Papernot, Patrick McDaniel, and Ian Goodfellow.
\newblock Transferability in machine learning: from phenomena to black-box
  attacks using adversarial samples.
\newblock {\em arXiv preprint arXiv:1605.07277}, 2016.

\bibitem{pinot2020randomization}
Rafael Pinot, Raphael Ettedgui, Geovani Rizk, Yann Chevaleyre, and Jamal Atif.
\newblock Randomization matters how to defend against strong adversarial
  attacks.
\newblock In {\em International Conference on Machine Learning}, pages
  7717--7727. PMLR, 2020.

\bibitem{drivingAdvML}
Adnan Qayyum, Muhammad Usama, Junaid Qadir, and Ala Al-Fuqaha.
\newblock Securing connected \& autonomous vehicles: Challenges posed by
  adversarial machine learning and the way forward.
\newblock {\em IEEE Communications Surveys \& Tutorials}, 22(2):998--1026,
  2020.

\bibitem{BaRT}
Edward Raff, Jared Sylvester, Steven Forsyth, and Mark McLean.
\newblock Barrage of random transforms for adversarially robust defense.
\newblock In {\em 2019 IEEE/CVF Conference on Computer Vision and Pattern
  Recognition (CVPR)}, pages 6521--6530, 2019.

\bibitem{transferSNN}
Nitin Rathi and Kaushik Roy.
\newblock Diet-snn: A low-latency spiking neural network with direct input
  encoding and leakage and threshold optimization.
\newblock {\em IEEE Transactions on Neural Networks and Learning Systems},
  2021.

\bibitem{rathi2021diet}
Nitin Rathi and Kaushik Roy.
\newblock Diet-snn: A low-latency spiking neural network with direct input
  encoding and leakage and threshold optimization.
\newblock {\em IEEE Transactions on Neural Networks and Learning Systems},
  2021.

\bibitem{sengupta2018mtdeep}
Sailik Sengupta, Tathagata Chakraborti, and Subbarao Kambhampati.
\newblock Mtdeep: boosting the security of deep neural nets against adversarial
  attacks with moving target defense.
\newblock In {\em Workshops at the thirty-second AAAI conference on artificial
  intelligence}, 2018.

\bibitem{demystifying}
Chawin Sitawarin, Zachary~J Golan-Strieb, and David Wagner.
\newblock Demystifying the adversarial robustness of random transformation
  defenses.
\newblock In {\em International Conference on Machine Learning}, pages
  20232--20252. PMLR, 2022.

\bibitem{tramer2020adaptive}
Florian Tramer, Nicholas Carlini, Wieland Brendel, and Aleksander Madry.
\newblock On adaptive attacks to adversarial example defenses.
\newblock {\em Advances in neural information processing systems}, (33), 2020.

\bibitem{wang2019improving}
Yisen Wang, Difan Zou, Jinfeng Yi, James Bailey, Xingjun Ma, and Quanquan Gu.
\newblock Improving adversarial robustness requires revisiting misclassified
  examples.
\newblock In {\em International Conference on Learning Representations}, 2019.

\bibitem{TiT}
Chang Xiao and Changxi Zheng.
\newblock One man's trash is another man's treasure: Resisting adversarial
  examples by adversarial examples.
\newblock In {\em Proceedings of the IEEE/CVF Conference on Computer Vision and
  Pattern Recognition}, pages 412--421, 2020.

\bibitem{xu2022securing}
Nuo Xu, Kaleel Mahmood, Haowen Fang, Ethan Rathbun, Caiwen Ding, and Wujie Wen.
\newblock Securing the spike: On the transferabilty and security of spiking
  neural networks to adversarial examples.
\newblock {\em arXiv preprint arXiv:2209.03358}, 2022.

\bibitem{zhang2019theoretically}
Hongyang Zhang, Yaodong Yu, Jiantao Jiao, Eric Xing, Laurent El~Ghaoui, and
  Michael Jordan.
\newblock Theoretically principled trade-off between robustness and accuracy.
\newblock In {\em International conference on machine learning}, pages
  7472--7482. PMLR, 2019.

\bibitem{FAT}
Jingfeng Zhang, Xilie Xu, Bo Han, Gang Niu, Lizhen Cui, Masashi Sugiyama, and
  Mohan Kankanhalli.
\newblock Attacks which do not kill training make adversarial learning
  stronger.
\newblock In {\em International conference on machine learning}, pages
  11278--11287. PMLR, 2020.

\end{thebibliography}
